# RobustDefect-LLM: Explainable and Robustness-Aware Industrial Surface Defect Classification with Decision Support and AI-Assisted Reporting

Nazlıcan Düşünmez, Halûk Gümüşkaya
Department of Computer Engineering,
İstanbul Arel University, 34537 Istanbul, Türkiye
Email: nazli.dsnmz@gmail.com, halukgumuskaya@arel.edu.tr

**Abstract**
This paper presents RobustDefect-LLM, an industrial surface-defect inspection framework integrating deep-learning classification, operator-facing visual evidence, confidence-aware decision support, controlled AI-assisted reporting, traceable storage, and mobile interaction in a unified quality-control workflow. Here, robustness-aware denotes explicit evaluation under controlled image degradation and confidence-aware review routing, not an intrinsic robustness guarantee. Four transfer-learning-based convolutional neural networks, ResNet50, EfficientNet-B0, DenseNet121, and MobileNetV3-Large, were evaluated on 1,799 images from the six-class NEU-DET dataset using fixed training, validation, and held-out in-domain test partitions. MobileNetV3-Large achieved the highest numerical test accuracy (99.26%) and macro F1-score (0.9926), with a bootstrap 95% accuracy CI of 0.9815-1.0000. An exact paired McNemar test found no significant difference from DenseNet121 ($p = 1.000$). The selected model averaged 0.060 s per CPU forward pass (16.66 FPS). Under combined synthetic degradation, accuracy fell to 87.78% at mild intensity and below 40% at stronger intensities, revealing sensitivity to severe image-quality deterioration. Grad-CAM supplied visual evidence, while predictions with confidence below 0.90 or a top-2 margin below 0.10 were routed to HUMAN REVIEW. This conservative policy provided 12.22% automatic coverage and 100% observed selective accuracy among 33 eligible cases (95% CI: 89.43%-100.00%), while routing both observed classification errors to review. Under nominal controlled conditions, all 100 generated reports passed deterministic consistency checks, with a mean latency of 1.66 s. Results support the feasibility of the integrated workflow while emphasizing the need for calibration, repeated evaluation, and real-world industrial validation.

## 1. Introduction

Industrial quality inspection is a fundamental component of modern manufacturing because surface defects can reduce product reliability, increase production costs, and compromise customer satisfaction. Conventional inspection processes often rely on human operators or handcrafted image-processing techniques. Although these approaches can perform well under controlled conditions, they remain sensitive to operator fatigue, subjectivity, illumination changes, surface texture variations, and image-quality degradation. As manufacturing systems become increasingly automated and production volumes continue to grow, intelligent computer vision systems have become essential for achieving consistent, reliable, and scalable quality inspection.

Deep learning has significantly improved the capability of automated inspection systems to recognize complex visual patterns. In particular, convolutional neural networks (CNNs) have demonstrated excellent performance in industrial surface defect classification by automatically learning hierarchical texture and structural features directly from image data. Despite these advances, many high-performing CNN-based systems still operate as black boxes, providing little information about how predictions are produced. In practical manufacturing environments, classification accuracy alone is insufficient. Quality engineers must also understand why a defect has been predicted, how reliable the prediction is, and whether the result should directly influence production decisions.

Consequently, practical deployment increasingly requires explainability, uncertainty awareness, traceability, and human-centered decision support in addition to high predictive performance.

To address these challenges, this study proposes RobustDefect-LLM, an integrated industrial inspection framework that combines defect classification, operator-oriented visual explanation, confidence-aware decision support, and controlled AI-assisted report generation. The framework employs transfer learning with four convolutional neural network (CNN) architectures—ResNet50, EfficientNet-B0, DenseNet121, and MobileNetV3-Large—to classify six steel surface defect categories from the publicly available NEU-DET dataset. Grad-CAM is employed to visualize class-discriminative image regions, while a deterministic decision engine converts prediction confidence and defect severity into structured quality-control recommendations. A full-stack implementation consisting of a FastAPI backend, a MongoDB database, and a React Native mobile application further demonstrates how explainable AI can be integrated into a practical inspection workflow that supports visualization, traceability, and operator-assisted decision making.

The primary contribution of this work is the integration of multiple complementary components into a unified industrial quality-control workflow. In addition to evaluating four transfer-learning-based CNN architectures on identical training, validation, and held-out in-domain test partitions under architecture-specific training configurations, the proposed framework integrates visual explanation, confidence-aware processing, deterministic decision logic, controlled AI-assisted reporting, persistent inspection records, and mobile interaction within a deployable software system. Furthermore, the study evaluates classification performance, analyzes robustness under controlled image perturbations, investigates confidence-aware decision making, and discusses practical deployment considerations for AI-assisted industrial inspection. In this study, robustness-aware refers to the explicit evaluation of performance under controlled image degradation and the use of confidence-aware review routing, rather than to an intrinsic robustness guarantee. Rather than focusing solely on maximizing benchmark accuracy, RobustDefect-LLM emphasizes trustworthy workflow integration by combining interpretable perception, operational decision support, and deployable software architecture within a single end-to-end inspection framework.

## 2. Related Work

Automated industrial surface inspection has progressed from handcrafted image-processing pipelines to learning-based models capable of representing complex texture and morphology. Traditional methods commonly rely on thresholding, filtering, edge extraction, morphology, or engineered texture descriptors. Although computationally efficient, these approaches remain sensitive to illumination changes, sensor noise, surface reflectance, and defect variability [1], [2], [3], [4]. Classical machine-learning methods partially improve adaptability by learning decision boundaries from manually designed features; however, their performance remains highly dependent on feature engineering and application-specific parameter tuning [1], [2], [4].

The emergence of deep learning fundamentally changed industrial inspection by enabling convolutional neural networks (CNNs) to learn hierarchical visual features directly from image data. The success of deep CNNs in large-scale image recognition [5] together with advances in deep representation learning [6] established the foundation for modern defect classification systems. ImageNet [7] further enabled transfer learning, allowing pretrained feature representations to be adapted to relatively small industrial datasets. Training techniques such as batch normalization [8] and established deep-learning optimization practices [9] further improved convergence stability and reproducibility, making transfer learning a widely adopted strategy for industrial surface inspection.

Recent industrial studies demonstrate substantial progress in supervised defect analysis. Chen et al. investigated deep-learning-based surface defect detection for industrial components [10]. He et al. surveyed deep-learning approaches for industrial surface inspection [1], while Ameri et al. systematically reviewed deep-learning methods for surface defect detection [2]. Liu et al. focused on real-time inspection systems and deployment-oriented constraints [3], and Prunella et al. presented a comprehensive review of industrial surface defect detection techniques [4]. Collectively, these studies report significant advances in classification, localization, detection, and segmentation. However, comparatively less attention has been devoted to integrating prediction, explanation, confidence-aware decision support, traceability, and reporting into a unified industrial inspection workflow.

Benchmark datasets have played an important role in evaluating industrial inspection methods under standardized conditions. The NEU Surface Defect Database (NEU-DET) [11], used in this study, contains six representative steel surface defect categories and has become a widely adopted benchmark for supervised surface defect classification. Nevertheless, benchmark accuracy alone does not establish robustness under changes in illumination, sensor noise, acquisition conditions, material appearance, or previously unseen defect patterns. These limitations motivate complementary robustness analysis together with explicit handling of uncertain predictions rather than assuming that every model prediction should directly determine an operational quality-control action.

Anomaly detection represents another important research direction for industrial quality inspection. Instead of requiring exhaustive labels for every defect category, anomaly-detection methods learn the distribution of normal samples and identify deviations from expected patterns. The review by Shukla et al. highlights the importance of robustness and generalization in industrial anomaly detection [12]. Although anomaly-detection approaches are valuable for identifying previously unseen defects, they generally do not provide defect-specific class labels, deterministic quality-control rules, structured reporting, or operator-oriented explainability. Therefore, supervised classification and anomaly detection should be regarded as complementary rather than competing approaches.

Explainable artificial intelligence has also become increasingly important for industrial decision support. Grad-CAM generates class-discriminative localization maps that indicate image regions contributing to CNN predictions [13]. Such visualizations allow operators to inspect whether the model focuses on plausible defect regions rather than irrelevant background textures. However, Grad-CAM does not provide causal explanations and should therefore be interpreted as operator-facing visual evidence rather than proof of model correctness. Similarly, large language models such as Llama-3.1-8B-Instant [14], accessed through the Groq API [15], can transform structured prediction outputs into readable inspection reports, but unrestricted text generation may reduce auditability. Consequently, controlled report generation constrained by structured prediction data is more appropriate for industrial quality-control applications.

The research gap addressed in this study is therefore not the absence of accurate classifiers, explainability methods, or language-generation models individually. Instead, the gap lies in integrating these capabilities into a unified industrial inspection workflow that combines defect classification, operator-facing visual explanation, confidence-aware decision support, deterministic quality-control logic, controlled AI-assisted reporting, traceability, and mobile interaction. RobustDefect-LLM addresses this system-level gap by comparatively evaluating four transfer-learning-based CNN architectures—ResNet50 [16], EfficientNet-B0, DenseNet121, and MobileNetV3-Large—and integrating Grad-CAM visualization, confidence-aware processing, rule-based decision support, persistent inspection records, and controlled report generation within a full-stack prototype. The primary contribution of this work therefore lies in the integration of these components into a practical and explainable industrial inspection system rather than in proposing a novel classification architecture. Table 1 summarizes representative industrial surface defect inspection approaches, their main advantages and limitations, and their relevance to the proposed RobustDefect-LLM framework.

## 3. Proposed RobustDefect-LLM Framework

The proposed framework is designed around four connected functions: perception, explainability, decision support, and reporting. The perception component receives a steel surface image and predicts the defect class using a convolutional neural network (CNN). The explainability component applies Grad-CAM to generate a heatmap highlighting the image regions that most strongly influence the model prediction. The decision-support component combines the predicted defect class, confidence score, top-2 confidence margin, and defect severity to produce structured quality-control recommendations while explicitly handling uncertain predictions. Finally, the reporting component generates a controlled inspection report that can be reviewed by an operator or quality engineer.

**Table 1.** Summary of representative industrial surface defect inspection approaches.

| Approach | Main advantage | Main limitation | Relevance to this work |
|---|---|---|---|
| **Traditional image processing [1], [2], [3], [4]** | Low computational cost and simple implementation | Sensitive to illumination, noise, and surface variation | Motivates the need for learned feature extraction |
| **Classical machine learning [1], [2], [4]** | Useful with small datasets and interpretable features | Requires handcrafted descriptors and tuning | Shows limitations of feature engineering |
| **CNN-based classification [1], [2], [3], [4], [5], [6]** | Automatic feature learning and high accuracy | Black-box predictions and limited explanation | Forms the foundation for comparative evaluation of modern CNN architectures |
| **Transfer learning [7], [9], [10], [17], [18]** | Improves performance with limited labeled data | Depends on pretrained features and dataset similarity | Used for the comparative evaluation of ResNet50, EfficientNet-B0, DenseNet121, and MobileNetV3-Large |
| **Explainable AI / Grad-CAM [13]** | Highlights influential image regions | Provides visual evidence but not complete causal reasoning | Integrated to provide operator-facing visual evidence during inspection |
| **Anomaly detection [12]** | Can identify deviations without exhaustive labels for every defect category | May not provide defect-specific class semantics or operational QC actions | Motivates future hybrid supervised and anomaly-detection extensions |
| **LLM-assisted reporting [14], [15]** | Generates readable inspection summaries from structured results | Must be constrained to avoid unsupported or inconsistent free-form output | Implemented as a controlled schema-based reporting layer |

## 3.1. Rule-based Decision

The rule-based decision engine used in RobustDefect-LLM is summarized in Algorithm 1. The algorithm combines the predicted defect class, confidence score, top-2 confidence margin, and predefined defect severity to generate structured quality-control recommendations.

**Algorithm 1** Rule-based decision engine

```
1:  Require:
2:     ŷ      Predicted defect class
3:     c      Confidence score (maximum softmax probability)
4:     Δ      Top-2 confidence margin (p_(1) − p_(2))
5:     s      Defect severity ∈ {Low, Medium, High}
6:  Ensure: Decision ∈ {HUMAN REVIEW, ACCEPT, REWORK, REJECT}
7:  if Δ < 0.10 then
8:     Decision ← HUMAN REVIEW
9:  else if c < 0.90 then
10:    Decision ← HUMAN REVIEW
11: else
12:    if s = Low then
13:       Decision ← ACCEPT
14:    else if s = Medium then
15:       Decision ← REWORK
16:    else {s = High}
17:       Decision ← REJECT
18:    end if
19: end if
20: return Decision
```

The prototype employed fixed operating thresholds of 0.90 for the maximum confidence score and 0.10 for the top-2 confidence margin. These thresholds were applied consistently throughout the experiments to implement a conservative decision policy. Accordingly, predictions with a confidence score below 0.90 or a top-2 confidence margin below 0.10 are routed to the HUMAN REVIEW pathway. Only predictions satisfying both criteria proceed to severity-based decisions, where low-,

medium-, and high-severity defects are assigned to ACCEPT, REWORK, and REJECT, respectively. The held-out in-domain test partition was reserved exclusively for final performance evaluation and was not used for model selection, threshold selection, or hyperparameter optimization. The empirical behavior of the selected thresholds is analyzed in Section 7.3. For production deployment, the threshold values should be recalibrated on representative validation data according to the desired trade-off between automatic coverage, review workload, and decision risk.

It should be noted that the NEU-DET dataset does not contain a defect-free (normal) class. Therefore, the ACCEPT recommendation does not indicate that the inspected product is free of defects. Instead, it indicates that the detected defect is classified as low severity with high prediction confidence, according to the predefined decision rules. Likewise, predictions assigned to HUMAN REVIEW represent uncertain model outputs requiring operator verification rather than automatic acceptance or rejection.

The defect severity used by the decision engine is not predicted by the CNN. Instead, it is assigned according to a predefined, configurable prototype severity mapping based on the predicted defect category. Specifically, scratches are assigned to the low-severity category; crazing, inclusion, and pitted surface are assigned to the medium-severity category; and patches together with rolled-in scale are assigned to the high-severity category. These severity assignments are implementation parameters and can be adapted to different industrial quality-control policies or application requirements.

The overall architecture is intentionally modular. Model inference, Grad-CAM generation, decision logic, report generation, data storage, and mobile visualization are separated into components that communicate through the backend API. This structure makes it easier to replace the model, improve the decision rules, add a no-defect category, or deploy the backend on a cloud or edge platform in future versions.

Figure 1 shows the main defect classification pipeline. The same workflow supports both captured images and uploaded images. The input is resized and normalized, the classifier generates top-class and top-2 probability outputs, Grad-CAM generates a visual explanation, and the decision engine produces a final ACCEPT, REWORK, REJECT, or HUMAN REVIEW action.

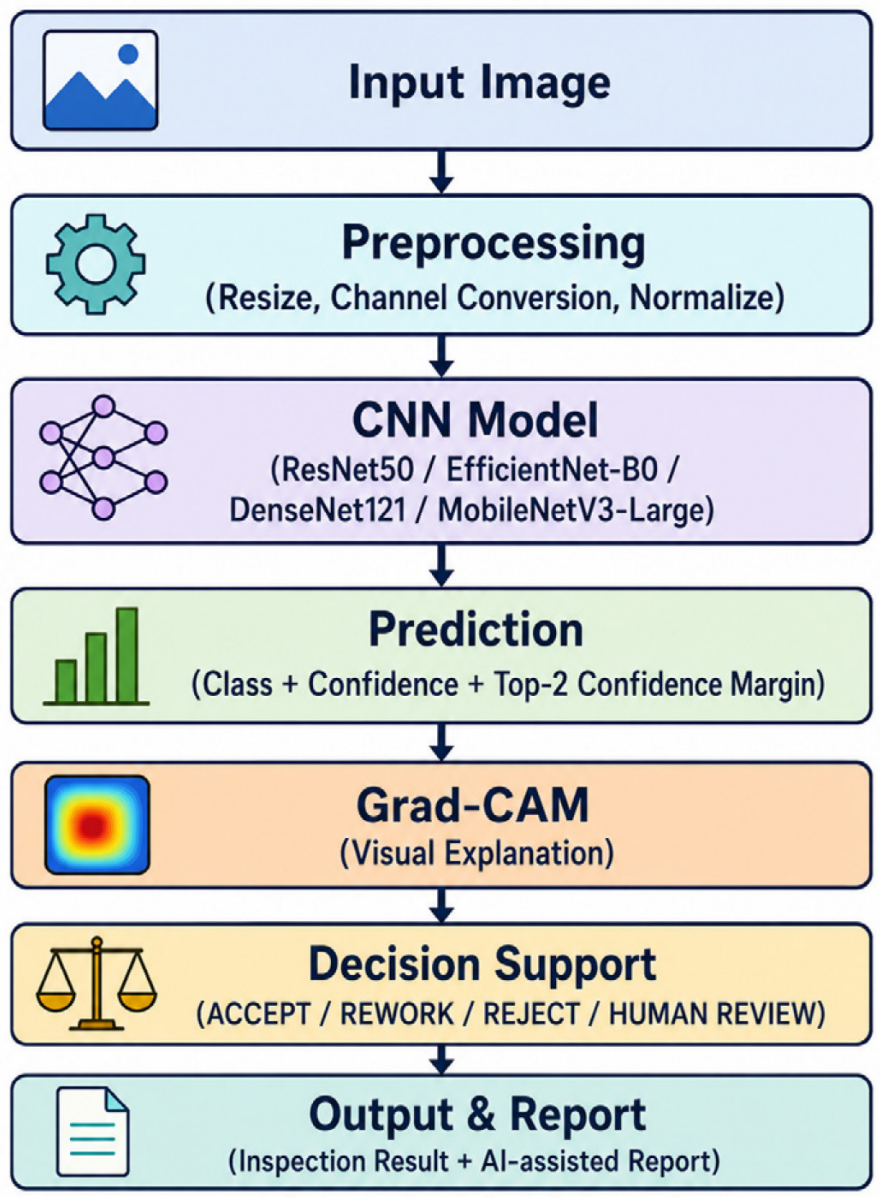

**Figure 1.** System architecture of the proposed defect classification pipeline.

## 3.2. Controlled AI-Assisted Reporting

The reporting module is designed as a controlled report-generation component rather than an unconstrained chatbot. The backend transmits only structured inspection fields—including the predicted defect class, confidence score, top-2 confidence margin, defect severity, decision recommendation, and selected metadata—to the LLM service. The LLM generates only a candidate textual report from this structured inspection record. The backend then validates the generated report against the structured inspection results and automatically replaces it with a deterministic rule-based fallback report if inconsistencies, missing required fields, or unsupported statements are detected. MongoDB stores structured inspection records, predictions, decisions, validated report content or metadata, and artifact paths. Raw uploaded images, Grad-CAM visualizations, and optional PDF/JSON report files are stored in the local file-storage layer. The validated report is also returned to the client in the JSON API response. A predefined prompt constrains the response to a fixed report format, reducing the possibility of irrelevant, inconsistent, or unsupported content. The implementation uses the Llama-3.1-8B-Instant model [14] through the Groq API [15] for optional AI-assisted report generation. A deterministic rule-based fallback report is generated automatically if report validation fails or the external reporting service is unavailable. This validation-and-fallback procedure preserves consistency, traceability, and auditability while ensuring that the LLM cannot alter the underlying classification or quality-control decision.

The reporting module is designed as an explanatory layer over structured prediction results rather than an independent decision maker. Consequently, the generated report does not modify the prediction or decision recommendation produced by the quality-control engine. Instead, it provides a standardized natural-language explanation to support inspectors during the review process. Human verification remains essential, particularly for predictions assigned to the HUMAN REVIEW pathway or for safety-critical quality-control decisions. The quantitative evaluation of the controlled reporting module is presented in Section 6. Table 2 summarizes the core components of the proposed RobustDefect-LLM framework together with their functions and implementation choices.

**Table 2.** Core components of the RobustDefect-LLM methodology.

| Component | Function in the system | Implementation choice |
|---|---|---|
| **Input image handling** | Accepts captured or uploaded surface images | Mobile application and FastAPI upload endpoint |
| **Preprocessing** | Resizes and normalizes images | 224 × 224 input, ImageNet normalization |
| **Primary classifier** | Predicts one of six defect classes | MobileNetV3-Large with transfer learning |
| **Comparative CNN models** | Provide comparative performance evaluation | ResNet50, EfficientNet-B0, DenseNet121, and MobileNetV3-Large |
| **Explainability** | Highlights influential image regions | Grad-CAM heatmap generation |
| **Decision engine** | Generates quality-control recommendations | Confidence score, top-2 confidence margin, severity rules, and HUMAN REVIEW pathway |
| **Reporting** | Generates structured inspection reports | Controlled LLM-assisted reporting (Groq API) |
| **Storage and history** | Supports traceability and review | MongoDB for structured records, predictions, decisions, validated report content or metadata, and artifact paths; local file storage for raw images, Grad-CAM visualizations, and generated PDF/JSON report files. |

## 3.3. Implementation-Aligned Mathematical Formulation

This subsection formalizes the transformations that are exposed by the implemented pipeline. It makes explicit how the preprocessing configuration reported in Section 4, the trained classifiers described in Section 5, and the decision, prototype, and reporting components described in Sections 7–9 convert one inspection image into an auditable record.

Let I be an acquired grayscale surface image. The implementation converts I to 3 channels, resizes it to 224 × 224 pixels, and applies ImageNet channel-wise normalization. For channel $\ell$, the model input is

$$x_\ell = \frac{\left[\mathrm{Resize}\left(\mathrm{Repeat}_3(I)\right)\right]_\ell - \mu_\ell}{\sigma_\ell}, \qquad \ell \in \{1,2,3\} \tag{1}$$

where $\mu_\ell$ and $\sigma_\ell$ are the ImageNet mean and standard deviation used by the pretrained backbone. Equation (1) is instantiated by the preprocessing choices in Table 3; consequently, training and inference must use the same channel conversion, resolution, and normalization parameters.

For $K = 6$ defect classes, the selected CNN f_θ produces a logit vector $z$ = f_θ(x). The deployed class probabilities are obtained with the softmax transformation

$$p_k(x) = \frac{\exp(z_k)}{\sum_{j=1}^{K} \exp(z_j)}, \qquad k = 1, \dots, K \tag{2}$$

and the predicted label, maximum softmax score, and top-2 margin are defined as

$$\hat{y} = \underset{k}{\operatorname{argmax}}\ p_k(x), \qquad c = p_{(1)}(x), \qquad \Delta = p_{(1)}(x) - p_{(2)}(x) \tag{3}$$

where $p_{(1)}$ and $p_{(2)}$ denote the largest and second-largest probabilities. The score $c$ is an operational confidence indicator, not a guaranteed probability of correctness; modern neural networks can be miscalibrated [19]. The margin Δ captures ambiguity between the two most likely defect classes and is the quantity reported as the top-2 confidence margin in Algorithm 1 and Table 13.

The HUMAN REVIEW branch is a selective prediction rule: the system abstains from an automatic quality-control action when either the maximum score or the class margin is below its configured threshold [20]. With the prototype settings $\tau_c = 0.90$ and $\tau_\Delta = 0.10$, the review indicator is

$$r(x) = \mathbb{1}[c < \tau_c \ \vee \ \Delta < \tau_\Delta] \tag{4}$$

where $\mathbb{1}[\cdot]$ is the indicator function. These thresholds are implementation parameters rather than universal constants. Their empirical behavior is examined in Section 7; before factory deployment they should be calibrated on representative production data and selected according to the acceptable trade-off between review workload and decision risk.

For visual evidence, Grad-CAM [13] uses the gradients of the predicted-class logit $z_{\hat{y}}(x)$ with respect to feature map $A^m$ in the selected convolutional layer. Defining the predicted-class score as $S_{\hat{y}}(x) = z_{\hat{y}}(x)$, and using spatial index (u, v) and $Z$ feature-map locations, the channel weight and heatmap are

$$\alpha_m^{\hat{y}} = \frac{1}{Z} \sum_u \sum_v \frac{\partial z_{\hat{y}}}{\partial A_{uv}^m}, \qquad L_{\mathrm{Grad-CAM}}^{\hat{y}} = \mathrm{ReLU}\left(\sum_m \alpha_m^{\hat{y}} A^m\right) \tag{5}$$

The heatmap is resized to the input-image coordinates and overlaid for operator review, as illustrated in Figures 6, 8, and 9. It indicates influential regions but does not validate the prediction or provide a causal explanation; therefore, it is retained as supporting evidence rather than used to override the classifier.

Let s ∈ {low, medium, high} be the configured defect-severity category supplied to the decision engine. The final recommendation implements Algorithm 1 as the following deterministic mapping:

$$d(x, s) = \begin{cases} \text{HUMAN REVIEW}, & r(x) = 1, \\ \text{ACCEPT}, & r(x) = 0 \ \wedge \ s = \text{low}, \\ \text{REWORK}, & r(x) = 0 \ \wedge \ s = \text{medium}, \\ \text{REJECT}, & r(x) = 0 \ \wedge \ s = \text{high}. \end{cases} \tag{6}$$

Because NEU-DET contains no defect-free class, ACCEPT means only that a confidently classified defect has low configured severity; it does not certify a defect-free product. The mapping keeps the operational decision traceable to $c$, Δ, $s$, and the two thresholds.

Finally, the reporting service receives the structured record $q = (\hat{y}, c, \Delta, s, d,$ metadata) and generates a candidate natural-language report $h(q)$. A validator $V$ checks that the required fields in the text agree with $q$. The report presented to the operator is

$$R(q) = \begin{cases} h(q), & V(h(q), q) = 1, \\ T(q), & \text{otherwise.} \end{cases} \tag{7}$$

where $T(q)$ is the deterministic fallback template. Thus, the LLM changes only the presentation of verified structured results; it cannot change the predicted class, severity, review status, or quality-control action. This formalization connects the controlled reporting strategy in Section 3.2 with the service interactions in Section 9 and the traceability and human-oversight requirements discussed in Sections 10 and 11.

## 4. Dataset and Preprocessing

The experiments were conducted using the publicly available NEU Surface Defect Database (NEU-DET) [11], a widely used benchmark for steel surface-defect classification. The dataset comprises grayscale images from six defect categories: crazing, inclusion, patches, pitted surface, rolled-in scale, and scratches. Representative samples are shown in Figure 2, highlighting the substantial differences in texture, morphology, spatial extent, and image contrast among the classes.

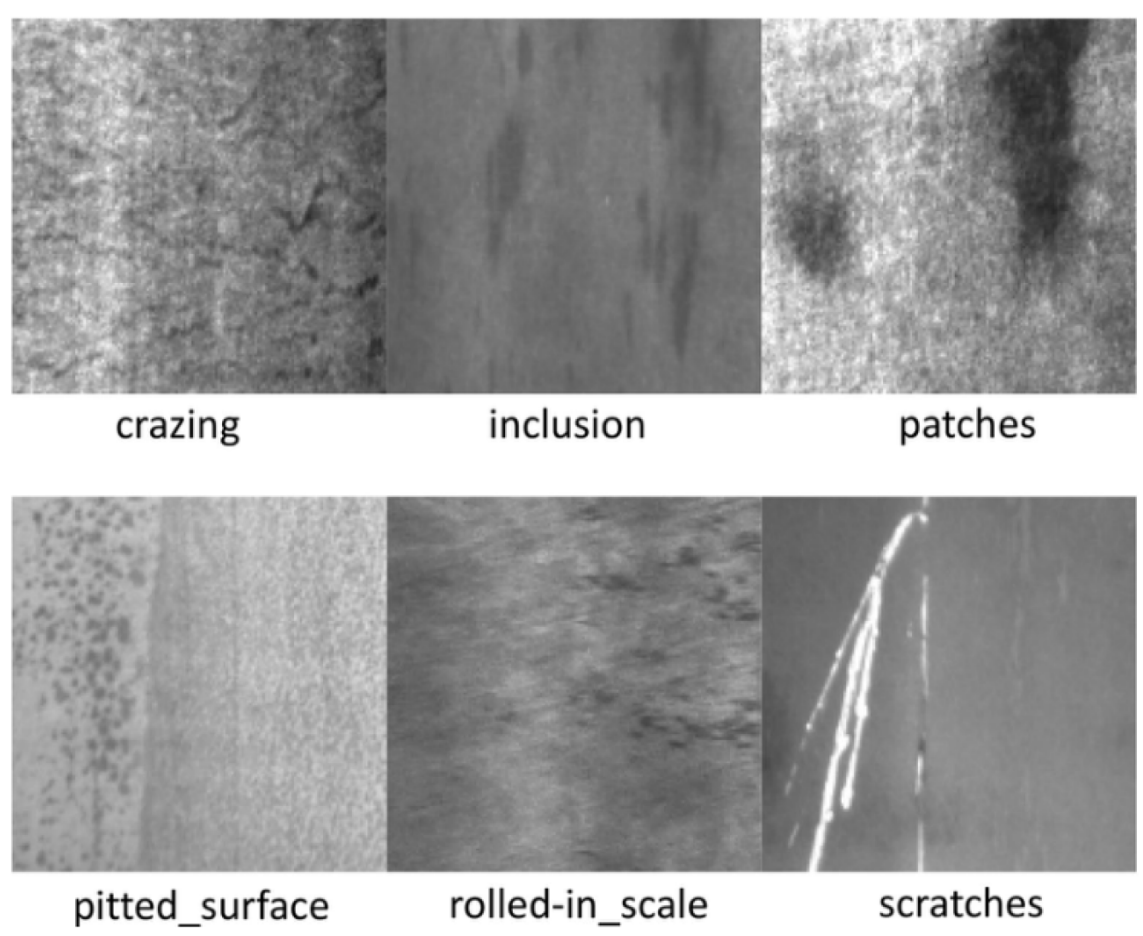

**Figure 2.** Representative defect classes from the NEU-DET dataset: crazing, inclusion, patches, pitted surface, rolled-in scale, and scratches.

The principal dataset characteristics and the preprocessing pipeline are summarized in Table 3. Each image was resized to 224 × 224 pixels, converted to the three-channel format required by the ImageNet-pretrained backbones, and normalized using ImageNet statistics.

**Table 3.** Dataset characteristics and preprocessing summary.

| Property | Value |
|---|---|
| **Dataset** | NEU-DET surface defect dataset |
| **Total samples** | 1,799 grayscale images |
| **Defect classes** | crazing, inclusion, patches, pitted surface, rolled-in scale, scratches |
| **Train / validation / test split** | 1259 / 270 / 270 images (approximately 70% / 15% / 15%) |
| **Random seed** | 42 |
| **Input resolution** | 224 × 224 pixels |
| **Preprocessing** | resize, channel conversion, ImageNet normalization |
| **Augmentation** | random flip, rotation, brightness/contrast jitter |
| **Main limitation** | limited dataset diversity and no explicit no-defect class |

During training, random horizontal flips, small random rotations, and brightness and contrast jitter were applied to increase sample variability and reduce overfitting. Saturation augmentation was intentionally omitted because the NEU-DET dataset consists of grayscale images that were converted to three channels only to satisfy the input requirements of the pretrained CNN backbones. These transformations were restricted to the training set; validation and test images underwent only the deterministic resizing, channel conversion, and normalization steps.

The original NEU-DET dataset is generally reported to contain 1,800 images, with 300 images per category. Verification of the downloaded files showed that one image was missing; consequently, the experiments used all 1,799 available images, with no additional exclusions made by the authors. The dataset was then divided into mutually exclusive training, validation, and test subsets using a fixed random seed of 42. No image was shared across the three subsets.

As reported in Table 4, the resulting split contained 1,259 training images, 270 validation images, and 270 test images, corresponding approximately to a 70%/15%/15% allocation. Five classes retained 300 samples, whereas the scratches class contained 299 samples; its training subset therefore included 209 rather than 210 images. The validation set was used for model selection and checkpoint selection, while the held-out in-domain test partition was reserved for the final evaluation. Because the original acquisition or production grouping could not be reconstructed, these results are treated as held-out in-domain evidence rather than sequence-disjoint or external-domain evidence.

**Table 4.** Class-wise distribution of the experimental dataset.

| Defect Class | Training | Validation | Test | Total |
|---|---|---|---|---|
| **Crazing** | 210 | 45 | 45 | 300 |
| **Inclusion** | 210 | 45 | 45 | 300 |
| **Patches** | 210 | 45 | 45 | 300 |
| **Pitted Surface** | 210 | 45 | 45 | 300 |
| **Rolled-in Scale** | 210 | 45 | 45 | 300 |
| **Scratches** | 209 | 45 | 45 | 299 |
| **Total** | 1,259 | 270 | 270 | 1,799 |

## 5. Model Training and Experimental Setup

Four transfer-learning-based convolutional neural network (CNN) architectures—ResNet50, EfficientNet-B0, DenseNet121, and MobileNetV3-Large—were selected to provide a balanced comparison across different network design philosophies. ResNet50 represents a well-established residual architecture [16], EfficientNet-B0 employs compound scaling to achieve an effective balance between accuracy and computational efficiency [17], DenseNet121 promotes feature reuse through dense connectivity [21], and MobileNetV3-Large is optimized for lightweight deployment while maintaining competitive classification performance [22]. Evaluating these representative architectures using the same train/validation/test split and evaluation protocol allows a consistent comparative evaluation under a common experimental setup.

All models were initialized with ImageNet-pretrained weights and fine-tuned using transfer learning. The classification layer of each network was replaced with a six-class output layer corresponding to the NEU-DET defect categories. Model training was performed using the Adam or AdamW optimizer [23], [24], depending on the architecture, together with early stopping to reduce overfitting and improve generalization. Table 5 summarizes the training configuration used for the evaluated CNN models. To quantify the statistical uncertainty of test performance, a non-parametric bootstrap analysis was performed on the held-out in-domain test partition. The test predictions were resampled with replacement 10,000 times using a random seed of 42. Accuracy, macro precision, macro recall, and macro F1-score were recomputed for each bootstrap sample, and the 2.5th and 97.5th percentiles of the empirical distributions were reported as 95% confidence intervals.

All training experiments were performed on a MacBook Air equipped with an Apple M1 processor and 8 GB of unified memory running macOS 15.5 (Build 24F74). The implementation used Python 3.14.3 and PyTorch 2.10.0 with the Apple Metal Performance Shaders (MPS) backend. Data loading was performed with num_workers = 0. The reported training durations include both the training and validation phases and correspond to a single training run for each architecture rather than an average over repeated runs. Early stopping was employed using the architecture-specific patience values reported in Table 5. The completed training epochs were 9 for EfficientNet-B0, 10 for ResNet50, 10 for DenseNet121, and 12 for MobileNetV3-Large. For each architecture, the checkpoint achieving the highest validation accuracy was retained for final evaluation (selected checkpoint epochs: 8, 9, 2, and 6, respectively). If multiple epochs achieved the same highest validation accuracy, the earliest occurrence was retained. The dataset split and bootstrap resampling both used a fixed random seed of 42. The same seed was also applied to the Python, NumPy, and PyTorch random-number generators. Consequently, model initialization (for the newly initialized classification layer), data-loader shuffling, stochastic data augmentation, and synthetic image degradations were all generated from seeded random-number generators. Each CNN architecture (EfficientNet-B0, ResNet50, DenseNet121, and MobileNetV3-Large) was trained once using this fixed experimental seed. Despite these controls, complete bitwise reproducibility cannot be guaranteed on the Apple Metal Performance Shaders (MPS) backend because certain low-level operations may remain nondeterministic. No initial backbone freezing was applied; all pretrained backbone parameters were fine-tuned throughout training.

**Table 5.** Training configuration of the evaluated CNN models.

| Parameter | EfficientNet-B0 | ResNet50 | DenseNet121 | MobileNetV3-Large |
|---|---|---|---|---|
| **Maximum epochs** | 12 | 10 | 10 | 12 |
| **Batch size** | 32 | 32 | 32 | 32 |
| **Learning rate** | 0.00005 | 0.0001 | 0.00005 | 0.00005 |
| **Optimizer** | AdamW | Adam | AdamW | AdamW |
| **Weight decay** | 5e-4 | 1e-4 | 5e-4 | 5e-4 |
| **Early stopping patience** | 3 | 4 | 4 | 4 |
| **Color jitter** | brightness=0.10, contrast=0.10 | brightness=0.15, contrast=0.15 | brightness=0.10, contrast=0.10 | brightness=0.10, contrast=0.10 |
| **Input size** | 224 × 224 | 224 × 224 | 224 × 224 | 224 × 224 |
| **Completed epochs** | 9 | 10 | 10 | 12 |
| **Selected checkpoint epoch** | 8 | 9 | 2 | 6 |
| **Learning-rate scheduler** | CosineAnnealingLR | CosineAnnealingLR | CosineAnnealingLR | CosineAnnealingLR |
| **Loss function** | CrossEntropyLoss | CrossEntropyLoss | CrossEntropyLoss | CrossEntropyLoss |
| **Fine-tuning strategy** | Full-network fine-tuning | Full-network fine-tuning | Full-network fine-tuning | Full-network fine-tuning |

Each architecture received a comparable exploratory hyperparameter tuning budget. A full Cartesian grid search was not performed; instead, a manually selected subset of candidate learning rates, optimizers, weight-decay values, early-stopping patience settings, and augmentation configurations was evaluated using the same training and validation protocol before the final held-out in-domain test evaluation. Exploratory tuning was terminated once additional manually tested configurations no longer produced meaningful improvements in validation accuracy or training stability. Once selected, the architecture-specific hyperparameter configurations were fixed and applied without further modification during the held-out in-domain test evaluation.

In addition to bootstrap-based uncertainty estimation, the paired classification performance of the two highest-performing models was statistically compared using an exact two-sided McNemar test. The test was conducted on the predictions obtained from the same held-out in-domain test samples, thereby accounting for the paired nature of the model outputs. Only discordant cases, in which one model produced the correct prediction while the other produced an incorrect prediction, contributed to the test. The statistical significance level was set to $\alpha = 0.05$.

The training behavior of the selected MobileNetV3-Large model is illustrated in Figure 3. Both training and validation losses decreased rapidly during the first few epochs and remained low thereafter, indicating stable convergence. Similarly, training and validation accuracy increased rapidly during the first few epochs, with validation accuracy reaching 95.19% by the third epoch and continuing to improve thereafter, ultimately reaching 100% at the selected checkpoint. The curves show stable convergence and no obvious train-validation divergence on the selected validation split.

The experimental evaluation employed accuracy, precision, recall, and macro F1-score as the primary performance metrics. Training and validation loss and accuracy curves of the selected MobileNetV3-Large deployment model were analyzed to assess convergence behavior and potential overfitting. In addition, confusion matrix analysis was performed on the test set to identify class-wise performance and recurring misclassification patterns.

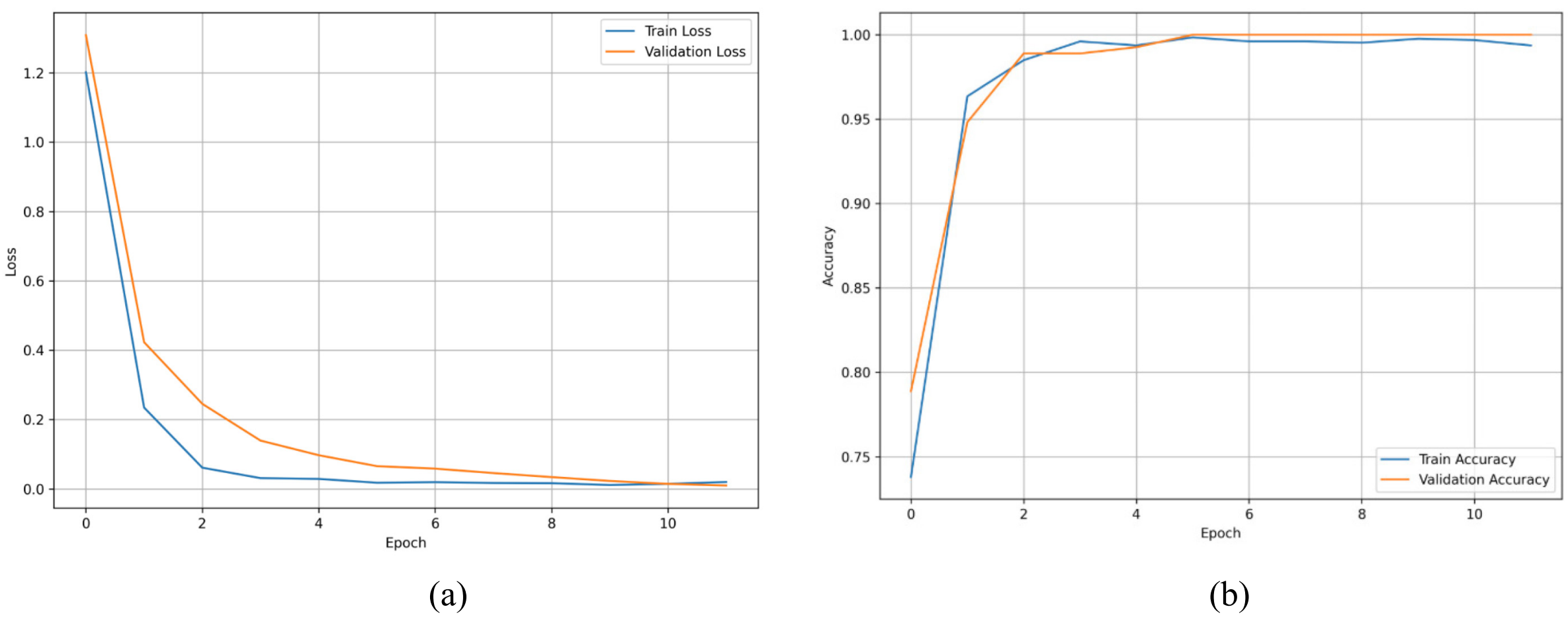

(a) (b)

**Figure 3.** Training performance of the selected MobileNetV3-Large deployment model: (a) training and validation loss curves and (b) training and validation accuracy curves.

## 6. Experimental Results

Four transfer-learning-based CNN architectures were evaluated using the same dataset split, input resolution, normalization procedure, and evaluation protocol, while architecture-specific training configurations are reported in Table 5. MobileNetV3-Large achieved the numerically highest test performance, followed closely by DenseNet121, whereas EfficientNet-B0 and ResNet50 produced lower results. As shown in Figure 3, the training and validation losses decreased rapidly during the initial epochs, while the corresponding accuracies increased and remained closely aligned. These curves show stable convergence and no obvious train–validation divergence on the selected validation split.

The maximum validation accuracy and total training time associated with the model-selection experiments are compared in Figure 4. MobileNetV3-Large and DenseNet121 both achieved a maximum validation accuracy of 100.00%, with total training times of 42.3 min and 56.8 min, respectively. EfficientNet-B0 achieved a maximum validation accuracy of 94.44% in 50.2 min, whereas ResNet50 achieved 80.74% in 44.4 min.

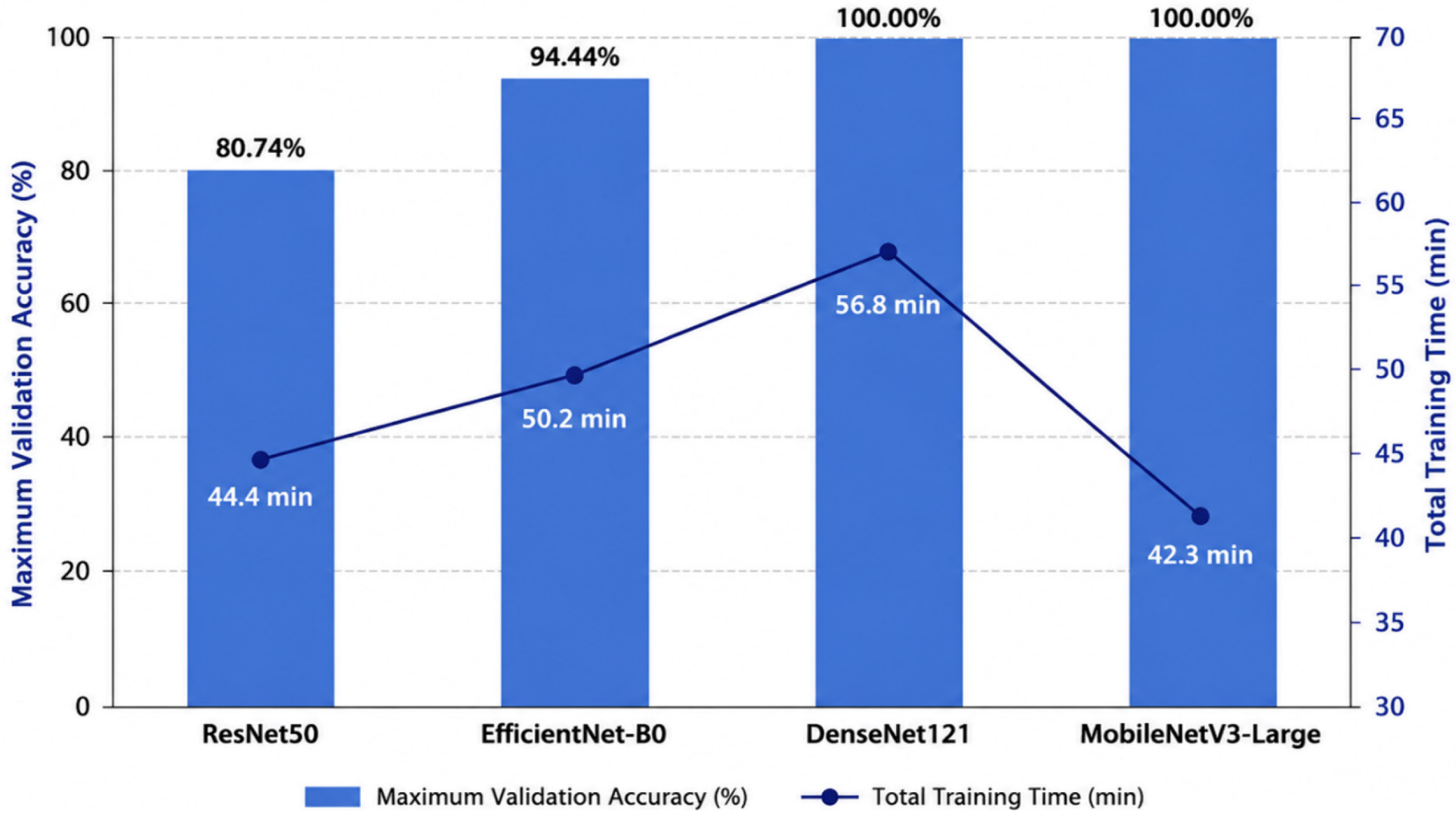

**Figure 4.** Maximum validation accuracy and total training time of the evaluated CNN models during model selection. Final quantitative performance on the held-out in-domain test partition is reported in Table 6.

The reported validation values correspond to the maximum validation accuracies recorded during the respective training runs, whereas the training times represent the total wall-clock duration of each run. These validation results, together with computational efficiency considerations, were used to select MobileNetV3-Large as the primary deployment model before final evaluation on the held-out in-domain test partition. Final quantitative performance on this test set is reported in Table 6.

Table 6 summarizes the final quantitative performance of all evaluated CNN architectures on the held-out in-domain test partition. MobileNetV3-Large achieved the numerically highest classification accuracy of 99.26%, together with a macro precision of 99.27%, a macro recall of 99.26%, and a macro F1-score of 99.26%. DenseNet121 also produced highly competitive results, achieving 98.89% accuracy, 98.93% macro precision, 98.89% macro recall, and 98.89% macro F1-score. EfficientNet-B0 achieved 95.19% accuracy, 95.63% macro precision, 95.19% macro recall, and 95.16% macro F1-score, whereas ResNet50 achieved 79.63% accuracy, 69.43% macro precision, 79.63% macro recall, and 73.24% macro F1-score. These findings indicate that MobileNetV3-Large provided the strongest observed balance between predictive performance and computational efficiency among the evaluated architectures.

**Table 6.** Final performance comparison on the held-out in-domain test partition.

| Model | Accuracy | Macro F1-score | Macro Precision | Macro Recall |
|---|---|---|---|---|
| **MobileNetV3-Large** | 0.9926 | 0.9926 | 0.9927 | 0.9926 |
| **DenseNet121** | 0.9889 | 0.9889 | 0.9893 | 0.9889 |
| **EfficientNet-B0** | 0.9519 | 0.9516 | 0.9563 | 0.9519 |
| **ResNet50** | 0.7963 | 0.7324 | 0.6943 | 0.7963 |

To quantify uncertainty associated with the finite held-out in-domain test sample, 95% bootstrap confidence intervals were computed from 10,000 bootstrap resamples, as summarized in Table 7. The two reported columns characterize uncertainty in overall classification accuracy and macro F1-score, respectively. MobileNetV3-Large achieved the highest point estimates for both accuracy and macro F1-score at 0.9926, with corresponding 95% confidence intervals of 0.9815–1.0000 and 0.9806–1.0000. DenseNet121 produced closely comparable results, with accuracy and macro F1-score values of 0.9889 and overlapping confidence intervals, supporting the conclusion that the two models perform similarly on the held-out in-domain test partition. EfficientNet-B0 achieved lower but still strong performance, whereas ResNet50 exhibited substantially lower point estimates and wider intervals, particularly for macro F1-score, indicating weaker and less balanced class-level performance. These intervals quantify test-sample uncertainty conditional on the trained models and observed test set; they do not capture variation arising from independent training runs, alternative data partitions, or model initialization.

**Table 7.** 95% bootstrap confidence intervals for the evaluated models on the held-out in-domain test partition.

| Model | Accuracy (95% CI) | Macro F1-score (95% CI) |
|---|---|---|
| **MobileNetV3-Large** | 0.9926 (0.9815–1.0000) | 0.9926 (0.9806–1.0000) |
| **DenseNet121** | 0.9889 (0.9741–1.0000) | 0.9889 (0.9746–1.0000) |
| **EfficientNet-B0** | 0.9519 (0.9259–0.9778) | 0.9516 (0.9241–0.9757) |
| **ResNet50** | 0.7963 (0.7481–0.8444) | 0.7324 (0.7018–0.7586) |

Because MobileNetV3-Large and DenseNet121 achieved closely comparable test performance, an exact two-sided McNemar test was used to determine whether the observed difference between their paired predictions was statistically significant. MobileNetV3-Large correctly classified 268 of the 270 test images, whereas DenseNet121 correctly classified 267 images. Both models correctly classified 265 samples. MobileNetV3-Large alone correctly classified three samples that were misclassified by DenseNet121, whereas DenseNet121 alone correctly classified two samples that were misclassified by MobileNetV3-Large. The exact McNemar test indicated that the difference was not statistically significant ($p = 1.000$). Therefore, although MobileNetV3-Large achieved the numerically highest accuracy, its classification performance was statistically comparable to that of DenseNet121 on the held-out in-domain test partition.

In addition to predictive performance, the computational complexity of the evaluated CNN architectures was analyzed in terms of the number of trainable parameters, floating-point operations (FLOPs), and model size. Computational complexity was estimated using the THOP library [25]. THOP returned multiply–accumulate (MAC) counts, which were converted to floating-point operation estimates by assuming that one MAC corresponds to two floating-point operations. Consequently, the computational complexity values reported in Table 8 are expressed as GFLOPs. As summarized in Table 8, MobileNetV3-Large achieved the highest classification performance while maintaining one of the lowest computational costs (0.467 GFLOPs). Although EfficientNet-B0 contains slightly fewer parameters, MobileNetV3-Large provides the best trade-off between predictive performance and computational efficiency for the proposed industrial inspection framework.

**Table 8.** Computational complexity comparison of the evaluated CNN architectures.

| Model | Parameters (M) | FLOPs (G) | Model Size (MB) |
|---|---|---|---|
| **EfficientNet-B0** | 4.015 | 0.828 | 15.70 |
| **ResNet50** | 23.520 | 8.263 | 90.04 |
| **DenseNet121** | 6.960 | 5.792 | 27.17 |
| **MobileNetV3-Large** | 4.210 | 0.467 | 16.31 |

To complement the computational complexity analysis, the deployment efficiency of the final MobileNetV3-Large model was evaluated using a CPU-based inference benchmark conducted on a MacBook Air equipped with an Apple M1 processor and 8 GB of unified memory. The benchmark environment consisted of macOS 15.5 (Build 24F74), Python 3.14.3, PyTorch 2.10.0, and torchvision 0.25.0. The benchmark was executed in CPU mode with a batch size of one, following ten warm-up iterations and one hundred timed forward-pass inference runs. As summarized in Table 9, MobileNetV3-Large achieved an average CNN forward-pass time of 0.060 s per image, corresponding to a throughput of 16.66 FPS, with approximately 213 MB of process memory usage. The reported measurements represent CNN forward-pass inference only and exclude image loading, preprocessing, Grad-CAM generation, API communication, database operations, mobile-interface rendering, and optional LLM-assisted report generation. Therefore, the reported inference time should not be interpreted as the complete end-to-end latency of the inspection system.

**Table 9.** CPU-only forward-pass inference benchmark of the deployed MobileNetV3-Large model.

| Metric | Value |
|---|---|
| **Average CNN forward-pass time** | 0.060 s/image |
| **Throughput** | 16.66 FPS |
| **Process memory usage** | ≈213 MB |

As shown in Figure 5, the selected MobileNetV3-Large model correctly classified 268 of the 270 held-out in-domain test images, corresponding to an overall accuracy of 99.26%. Perfect classification was obtained for the crazing, patches, pitted surface, and rolled-in-scale classes, with all 45 samples in each class assigned correctly. Only two errors were observed: one inclusion sample was misclassified as pitted surface, and one scratches sample was misclassified as inclusion. Consequently, the inclusion and scratches classes each achieved a recall of 44/45, or approximately 97.78%. The strong concentration of values along the main diagonal indicates highly consistent class discrimination, while the two off-diagonal errors suggest limited confusion between defect categories with partially similar local texture patterns.

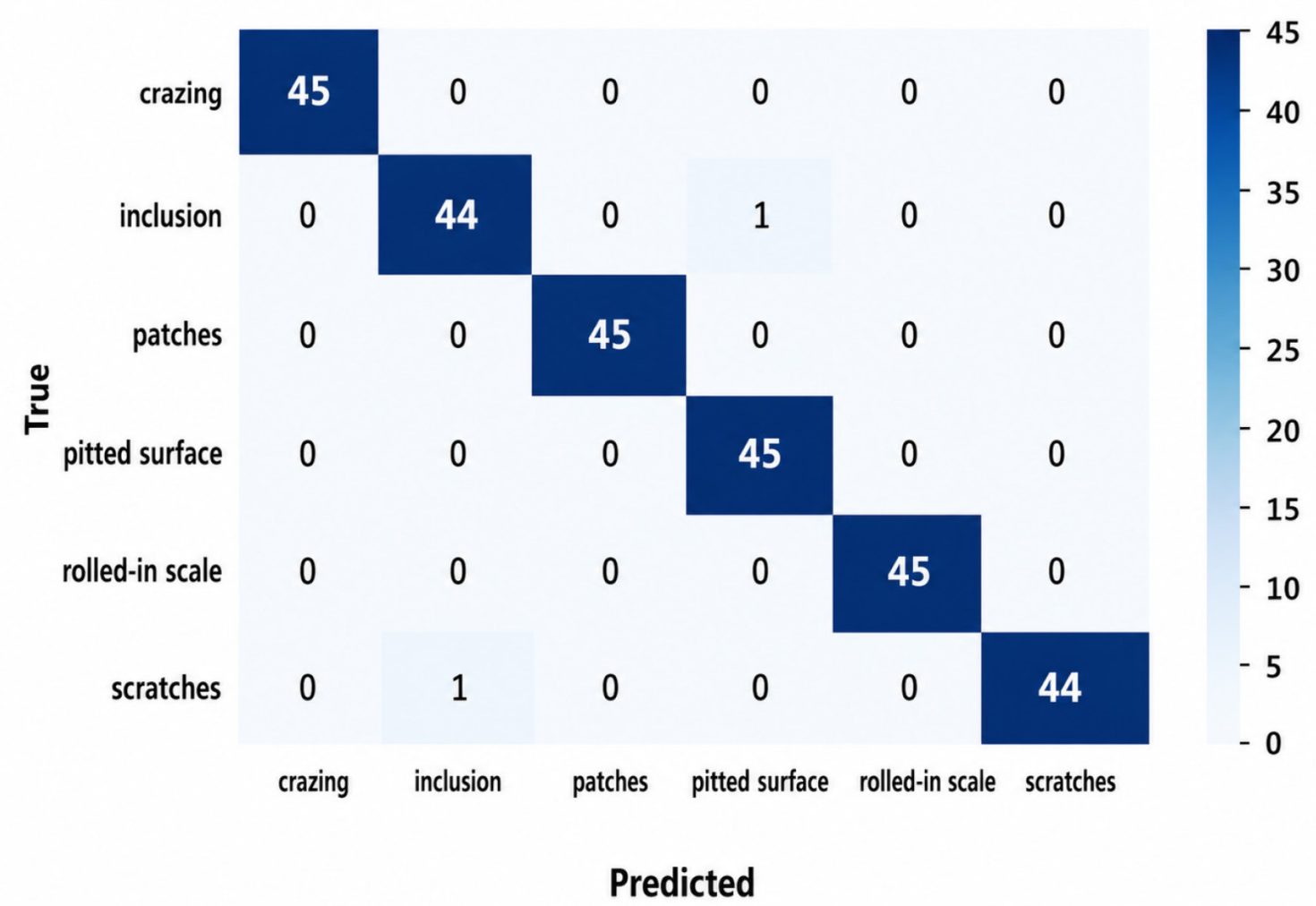

**Figure 5.** Confusion matrix of the selected MobileNetV3-Large model on the held-out in-domain test partition.

To quantitatively evaluate the nominal deterministic consistency of the controlled LLM reporting module, 100 controlled evaluation cases covering all six defect classes, multiple confidence ranges, severity categories, and quality-control decisions were evaluated under nominal operating conditions. As summarized in Table 10, each LLM-generated report was automatically checked using deterministic consistency rules that verified report completeness and agreement with the structured inspection results, including the predicted defect class, severity level, decision recommendation, and confidence interpretation. All 100 generated reports satisfied the predefined validation criteria without requiring fallback generation or experiencing API failures, resulting in a 100% validation pass rate under these controlled nominal-condition validation settings. The mean report-generation latency was 1.66 s per report.

**Table 10.** Quantitative validation of the controlled LLM reporting module under nominal operating conditions.

| Metric | Value |
|---|---|
| **Evaluation cases** | 100 |
| **Successfully validated reports** | 100 |
| **Validation pass rate** | 100.00% |
| **Fallback reports** | 0 |
| **Fallback rate** | 0.00% |
| **LLM API failures** | 0 |
| **Mean report generation latency** | 1.66 s |

As illustrated in Figure 6, the MobileNetV3-Large model classified the input image as a scratches defect with a confidence score of 77.51%. Because the confidence score is below the operational threshold of 0.90, the prediction is routed to the HUMAN REVIEW pathway rather than being converted into an automatic quality-control action. Although the scratches class is assigned a low severity under the prototype decision policy, severity-based actions are applied only when the predefined confidence and top-2 confidence margin requirements are satisfied. The corresponding Grad-CAM visualization provides operator-facing visual evidence of the image regions influencing the prediction. In this example, the activation pattern is concentrated around the visually salient linear defect region, while the prediction confidence remains below the threshold required for automatic decision support. The heatmap should not be interpreted as a causal explanation or as proof that the prediction is correct.

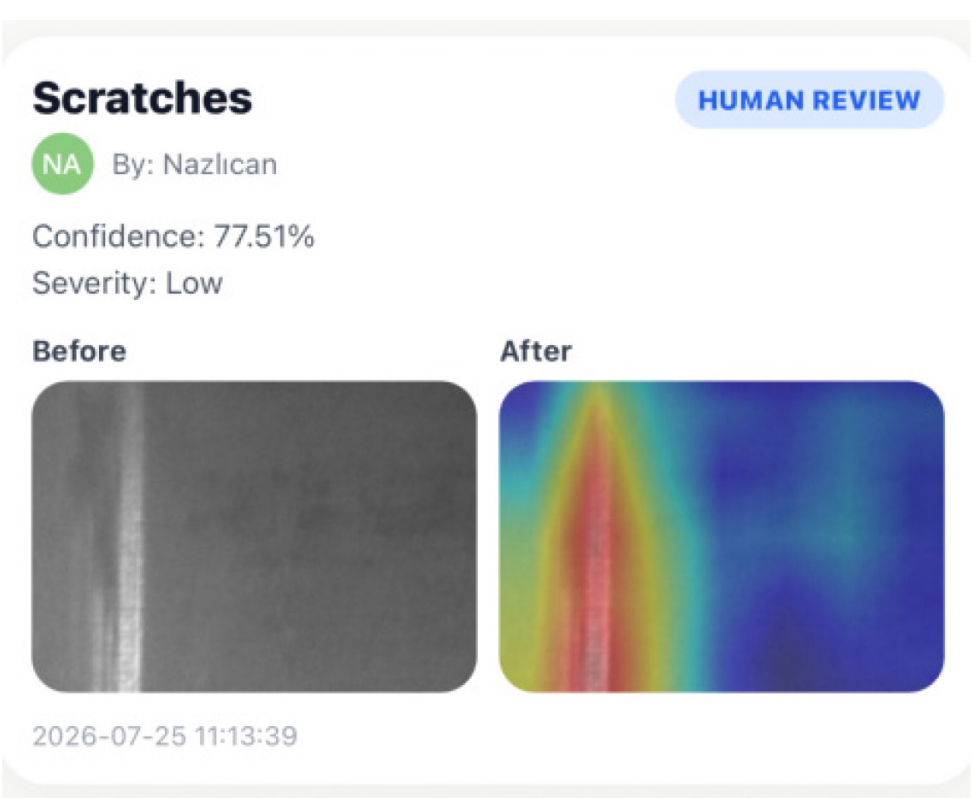

**Figure 6.** Example Grad-CAM visualization and HUMAN REVIEW output for a scratches prediction with confidence below the operational threshold.

# 7. Robustness, Error, and Confidence Analysis

## 7.1. Robustness under Synthetic Image Degradation

The robustness of the selected MobileNetV3-Large model was evaluated under progressively increasing synthetic image degradation using the held-out in-domain test partition. Four perturbation intensities were considered: 0.0, 0.3, 0.6, and 0.9. At each nonzero intensity level, Gaussian blur, brightness variation, and additive Gaussian noise were independently applied to each image with a probability of 0.70.

Gaussian blur was implemented using the PIL `GaussianBlur` operator with a radius of 2λ, where λ denotes the perturbation intensity. Brightness variation was generated using a multiplicative factor uniformly sampled from [1−0.4λ, 1+0.4λ], whereas zero-mean Gaussian noise with a standard deviation of 0.2λ was added after resizing the image to 224×224 pixels and converting it to tensor format. Pixel values were clipped to the valid [0,1] range before ImageNet normalization. Accordingly, perturbation intensities of 0.3, 0.6, and 0.9 corresponded to Gaussian-blur radii of 0.6, 1.2, and 1.8; brightness-factor ranges of [0.88,1.12], [0.76,1.24], and [0.64,1.36]; and Gaussian-noise standard deviations of 0.06, 0.12, and 0.18, respectively. No synthetic corruption was applied at intensity 0.0.

As summarized in Table 11, MobileNetV3-Large achieved 99.26% accuracy and a macro F1-score of 99.26% under clean-image conditions. At perturbation intensity 0.3, accuracy decreased to 87.78%, with a macro F1-score of 88.25%. Under more severe degradation, accuracy declined to 38.52% at intensity 0.6 and 33.70% at intensity 0.9.

**Table 11.** Robustness evaluation of MobileNetV3-Large under increasing synthetic image perturbation levels.

| Perturbation Intensity | Accuracy | Macro Precision | Macro Recall | Macro F1-score |
|---|---|---|---|---|
| **0.0** | 99.26 | 99.27 | 99.26 | 99.26 |
| **0.3** | 87.78 | 89.51 | 87.78 | 88.25 |
| **0.6** | 38.52 | 73.51 | 38.52 | 37.13 |
| **0.9** | 33.70 | 75.35 | 33.70 | 30.56 |

The increasing difference between macro precision and macro recall at the two highest perturbation levels suggests that severe degradation caused uneven class-wise prediction behavior rather than a uniform reduction across all defect categories. Overall, the model maintained high performance under clean conditions. Under the mild combined-corruption condition at intensity 0.3, accuracy decreased to 87.78%; whether this performance is operationally acceptable requires application-specific validation based on the costs of false acceptance, false rejection, and quality escapes. Performance deteriorated sharply at intensities 0.6 and 0.9, indicating substantial sensitivity to severe image degradation. The experiment should therefore be interpreted as a robustness stress test rather than evidence of factory-level robustness.

### 7.2. Classification Error Analysis

The confusion matrix in Figure 5 shows that MobileNetV3-Large correctly classified 268 of the 270 held-out in-domain test images. Perfect classification was achieved for the crazing, patches, pitted surface, and rolled-in-scale categories. Only two misclassifications were observed: one scratches sample was predicted as inclusion, and one inclusion sample was predicted as pitted surface. These isolated errors occurred between categories containing partially overlapping local textures, weak contrast differences, and similar elongated or localized defect patterns.

To examine the visual characteristics associated with these errors, Figure 7 presents the two misclassified test samples together with one additional correctly classified but visually challenging example. The elongated structures in the first two images exhibit characteristics that may be associated with both scratches and inclusion defects, whereas the localized dark structure in the third image shares textural similarities with inclusion and pitted-surface defects. These examples indicate that the remaining errors arise mainly from subtle defect boundaries, low contrast, and local texture overlap rather than broad confusion across all six categories.

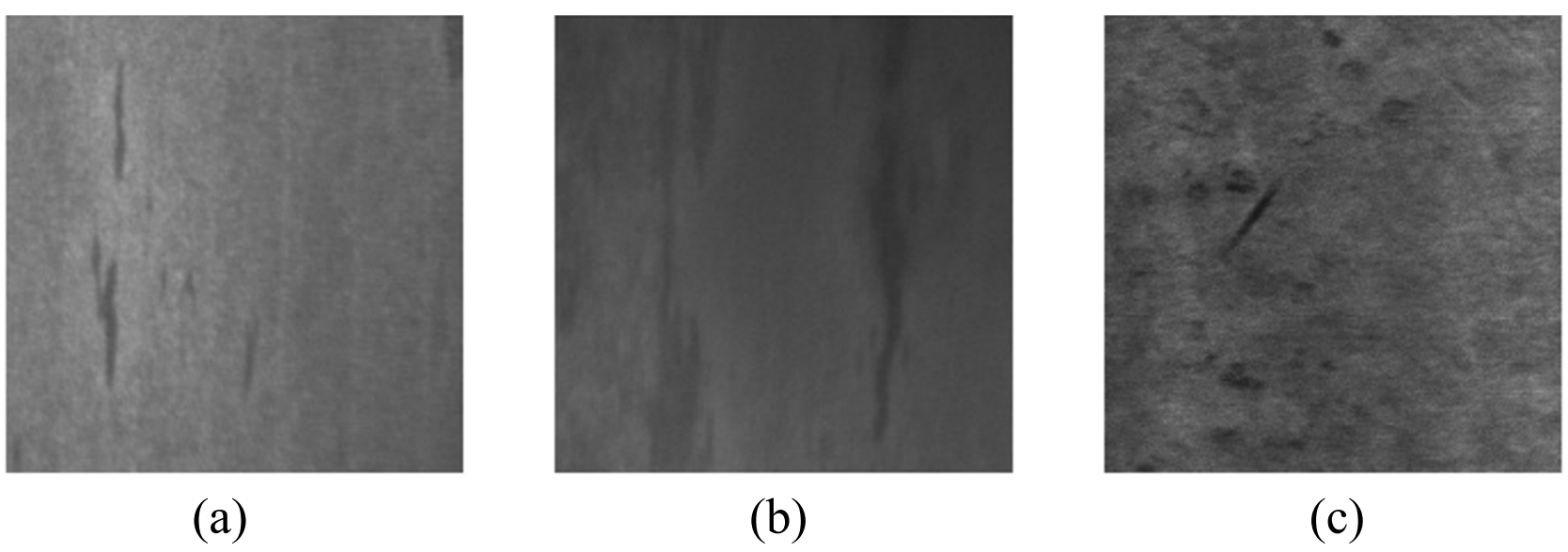

(a) (b) (c)

**Figure 7.** Representative visually challenging NEU-DET test samples used in the qualitative error analysis: (a) a scratches sample misclassified as inclusion, (b) an inclusion sample misclassified as pitted surface, and (c) an additional challenging sample exhibiting weak contrast and locally ambiguous texture.

The observed error types, their likely causes, and potential improvement strategies are summarized in Table 12. Additional class-specific training samples and stronger augmentation may improve discrimination between scratches and inclusion defects, whereas hard-example mining, improved feature learning, or localization-aware training may reduce confusion between inclusion and pitted-surface samples.

**Table 12.** Observed classification errors and potential improvement strategies.

| Error type | Observed frequency | Likely cause | Suggested improvement |
|---|---|---|---|
| **Scratches → Inclusion** | 1 sample | Visual similarity and thin linear patterns | More class-specific data and targeted augmentation |
| **Inclusion → Pitted surface** | 1 sample | Overlapping local texture patterns | Hard-example mining and improved feature learning |

### 7.3. Confidence-Aware Selective Decision Support

In addition to class prediction, the proposed decision-support mechanism evaluates the maximum softmax confidence score and the top-2 confidence margin. Rather than directly converting every prediction into an operational quality-control action, the system routes uncertain or ambiguous cases to HUMAN REVIEW. Under the final policy defined in Algorithm 1 and Equation (4), HUMAN REVIEW is triggered when $c < 0.90$ or $\Delta < 0.10$, where $c$ is the maximum confidence score and $\Delta$ is the difference between the highest and second-highest class probabilities. Only predictions satisfying both $c \geq 0.90$ and $\Delta \geq 0.10$ are eligible for the severity-based ACCEPT, REWORK, or REJECT decision rules. The operating thresholds ($c = 0.90$ and $\Delta = 0.10$) were selected empirically during prototype development to implement a conservative decision policy that prioritizes routing uncertain predictions to the HUMAN REVIEW pathway rather than maximizing automatic coverage. The thresholds were fixed before the held-out in-domain test evaluation and applied consistently throughout all experiments. No systematic threshold optimization or exhaustive grid search was performed. Accordingly, these values should be regarded as application-dependent prototype operating parameters rather than universally optimal industrial thresholds. Future work will investigate probability calibration, validation-based threshold optimization, and risk–coverage analysis for different industrial operating conditions.

To characterize prediction reliability in greater detail, a post-hoc confidence analysis was performed using three confidence ranges, as reported in Table 13. The boundary of 0.60 was used only to distinguish a very-low-confidence subset for analytical purposes; it does not independently determine the final operational action. The operational confidence threshold remained 0.90.

As shown in Table 13, 237 of the 270 test samples, corresponding to 87.78%, had confidence scores below the operational threshold of 0.90 and were therefore routed to HUMAN REVIEW. The remaining 33 samples, representing 12.22% of the test set, satisfied the confidence requirement. All 33 also had top-2 confidence margins of at least 0.10 and were therefore eligible for severity-based automated decisions.

**Table 13.** Confidence-based analysis of prediction reliability using the proposed decision-support strategy.

| Confidence Range | Number of Samples | Percentage (%) | Average Confidence (%) | Accuracy (%) | Decision Rule |
|---|---|---|---|---|---|
| **$c < 0.60$** | 14 | 5.19 | 50.61 | 85.71 | HUMAN REVIEW |
| **$0.60 \leq c < 0.90$** | 223 | 82.59 | 79.92 | 100.00 | HUMAN REVIEW |
| **$c \geq 0.90$** | 33 | 12.22 | 93.10 | 100.00 | Eligible for severity-based decision if $\Delta \geq 0.10$ |

Four test samples, corresponding to 1.48% of the test set, had top-2 confidence margins below 0.10. However, all four were already included among the 237 predictions with confidence below 0.90; consequently, the margin criterion did not introduce any additional HUMAN REVIEW cases in this experiment. Both observed classification errors occurred in the $c < 0.60$ group, whereas all predictions in the $0.60 \leq c < 0.90$ and $c \geq 0.90$ groups were classified correctly. The final selective-decision results can therefore be summarized as follows:

- automatic coverage: 12.22%;
- HUMAN REVIEW rate: 87.78%;
- selective accuracy among automatically eligible samples: 100.00% (33/33; exact 95% Clopper–Pearson CI: 89.43%–100.00%);
- observed selective risk: 0.00% (0/33);
- classification errors captured by HUMAN REVIEW: 2 of 2.

These findings demonstrate that the selected policy prevented the two observed errors from directly producing automated quality-control actions. However, although the observed selective accuracy was 100%, the exact 95% confidence interval (89.43%–100.00%) reflects the statistical uncertainty associated with the relatively small number of automatically eligible samples (33 cases). Nevertheless, the low automatic coverage indicates that the current thresholds are highly conservative. The large number of correct predictions in the $0.60 \leq c < 0.90$ range also suggests that the model's confidence scores may be undercalibrated. Future deployment should therefore include probability calibration and more comprehensive risk–coverage analyses to further refine the selected operating thresholds for different industrial operating conditions and application-specific risk requirements. The operating thresholds ($c = 0.90$ and $\Delta = 0.10$) were predefined conservative operating parameters of the prototype and were applied consistently throughout all experiments. They should therefore be regarded as application-dependent operating parameters rather than universally applicable industrial values. Different manufacturing environments may require recalibration using representative validation data to achieve an appropriate balance between review workload and operational risk.

Taken together, the robustness, qualitative-error, and confidence analyses reveal both the strengths and limitations of the proposed framework. MobileNetV3-Large achieves very high classification performance on the held-out in-domain test partition, while the HUMAN REVIEW pathway successfully captures the observed low-confidence errors. However, the substantial performance reduction under stronger image corruption and the conservative 12.22% automatic coverage demonstrate that further calibration, repeated robustness testing, broader datasets, and real production validation are necessary before autonomous industrial deployment.

## 8. System Prototype and Program User Interfaces

The practical contribution of RobustDefect-LLM is demonstrated through a full-stack industrial inspection prototype. The deep-learning inference engine employs the selected MobileNetV3-Large deployment model, implemented in PyTorch [26]. The backend is developed using FastAPI [27]. MongoDB [28] stores structured inspection records, predictions, decisions, validated report content or metadata, and artifact paths. Raw uploaded images, Grad-CAM visualizations, and generated PDF/JSON report files are stored in the local file-storage layer. The mobile client is implemented with React Native and Expo [29]. The backend provides RESTful endpoints for image upload, model inference, Grad-CAM generation, confidence estimation, decision support, report generation, history retrieval, and analytics. This modular architecture enables independent updates of the perception model, explainability module, decision-support layer, reporting service, and mobile application.

The mobile user interface was designed for inspectors and non-technical operators. The application provides image capture, image upload, prediction history, inspection reports, Grad-CAM visualization, and analytics dashboards. For each inspection, the interface displays the predicted defect class, confidence score, severity level, decision recommendation, and the corresponding Grad-CAM heatmap. This information allows operators to inspect the prediction together with its confidence and Grad-CAM evidence before making a quality-control decision. Figure 8 presents the prediction history together with the original defect image and the corresponding Grad-CAM visualization. Figure 9 presents the main user interfaces of the developed mobile application, including user authentication, image acquisition, confidence-aware defect analysis, quality-control decision presentation, and inspection history. These interfaces demonstrate how the proposed framework supports practical industrial quality-control workflows beyond standalone image classification. The main implementation and deployment characteristics of the developed prototype are summarized in Table 14.

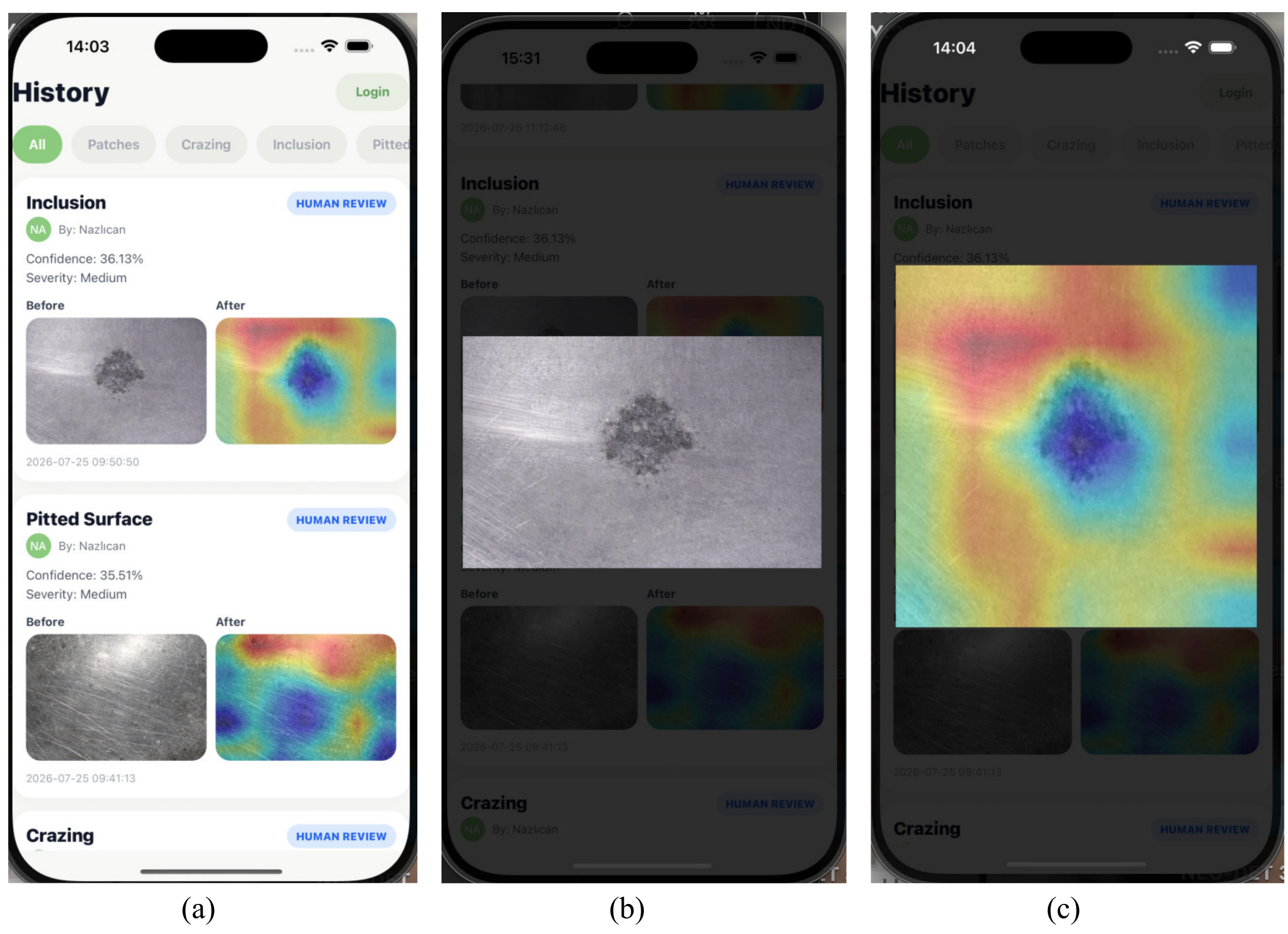

(a) (b) (c)

**Figure 8.** Mobile interface for confidence-aware defect inspection: (a) prediction history with HUMAN REVIEW decisions, (b) original defect image, and (c) corresponding Grad-CAM visualization.

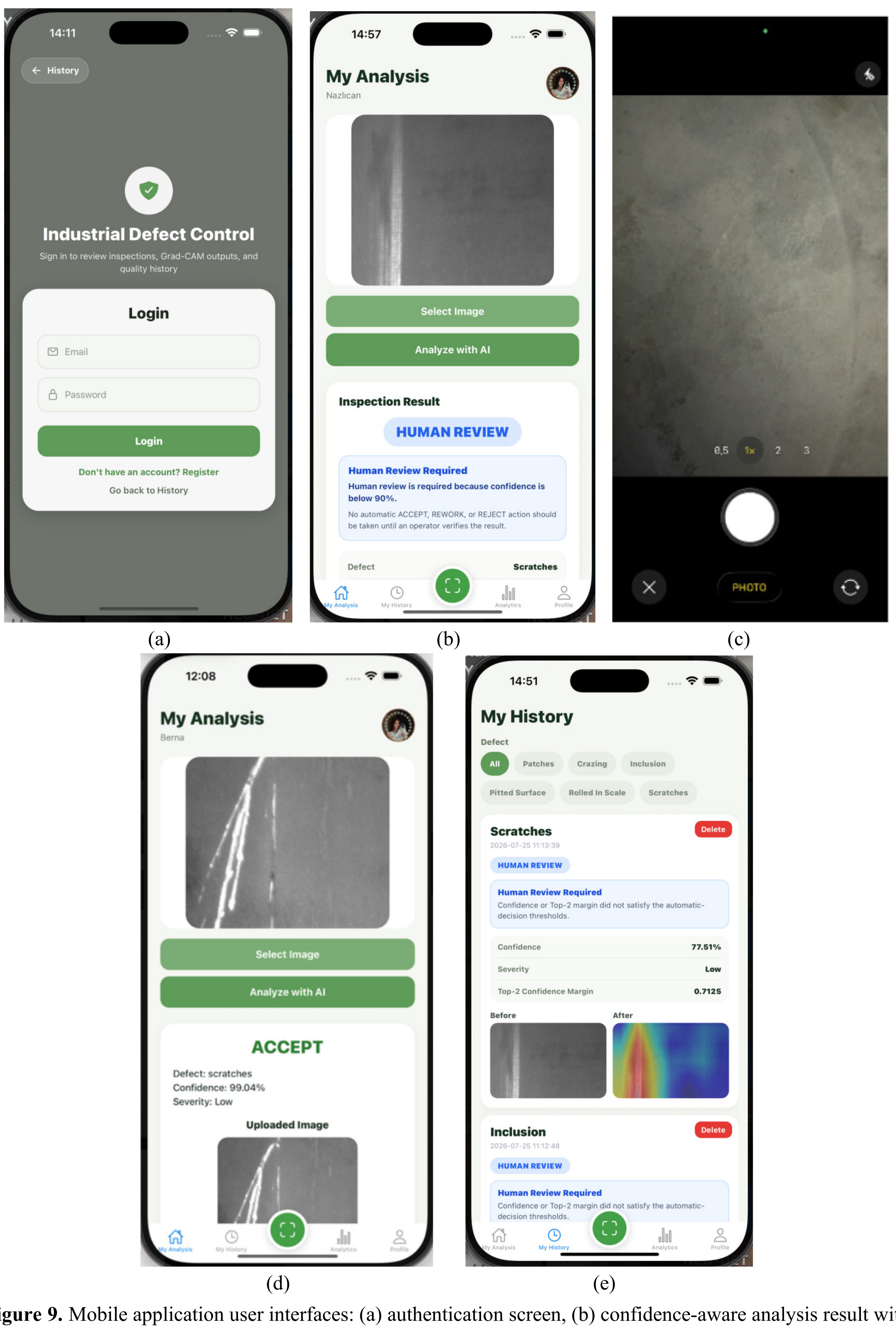

**Figure 9.** Mobile application user interfaces: (a) authentication screen, (b) confidence-aware analysis result with a HUMAN REVIEW decision, (c) camera-based image acquisition, (d) high-confidence analysis result with an automatic ACCEPT decision, and (e) inspection history showing the predicted defect class, confidence score, severity level, top-2 confidence margin, Grad-CAM visualization, and HUMAN REVIEW information.

Table 14. Implementation and deployment summary.

| Component | Implementation details |
|---|---|
| **Application type** | Mobile inspection application |
| **Frontend** | React Native with Expo |
| **Backend** | FastAPI Python server |
| **Model layer** | MobileNetV3-Large (primary model), DenseNet121, EfficientNet-B0, and ResNet50 implemented in PyTorch |
| **Storage** | MongoDB for structured inspection records, predictions, decisions, validated report content or metadata, and artifact paths; local file storage for raw uploaded images, Grad-CAM visualizations, and generated PDF/JSON report files |
| **CNN inference benchmark** | 0.060 s/image (16.66 FPS, batch size = 1) |
| **Deployment mode** | local machine backend with client-server architecture |
| **Key limitation** | not yet optimized for large-scale concurrent industrial deployment |

The average CNN inference time of 0.060 s per image reported in Table 14 was measured using the deployed MobileNetV3-Large model with a batch size of one, following ten warm-up iterations and one hundred timed inference runs. This benchmark represents CNN forward-pass inference only and excludes image loading, preprocessing, Grad-CAM generation, API communication, database operations, mobile-interface rendering, and optional LLM-assisted report generation. Therefore, the reported value should not be interpreted as the complete end-to-end response time of the mobile inspection system. The average latency of 1.66 s reported in Table 10 corresponds only to the optional LLM-assisted report-generation stage.

## 9. System Architecture and Interaction Model

The proposed system architecture separates the mobile client, API server, AI inference engine, decision-support module, report generation service, and database layer into independent components. This modular design improves maintainability, scalability, and flexibility by allowing each component to be developed, tested, or upgraded independently. For example, the MobileNetV3-Large classifier can be replaced by another classification, detection, or segmentation model without requiring changes to the mobile application, reporting module, or database infrastructure. Similarly, the locally deployed MongoDB database can be migrated to a cloud-based deployment to support large-scale and multi-site industrial inspection.

The detailed service architecture shown in Figure 10 illustrates how the AI inference engine, Grad-CAM explainability module, confidence analysis, decision-support logic, reporting service, API endpoints, and database components interact within the full-stack framework. Figure 11 presents the interaction sequence between the mobile application, backend server, AI inference engine, decision-support module, MongoDB database, and reporting service, demonstrating the complete workflow from image upload to prediction, explanation, decision recommendation, report generation, persistent storage, and result visualization. In this interaction, the API carries the structured record q defined in Section 3.3, and the reporting service applies the validation-and-fallback rule defined in Equation (7).

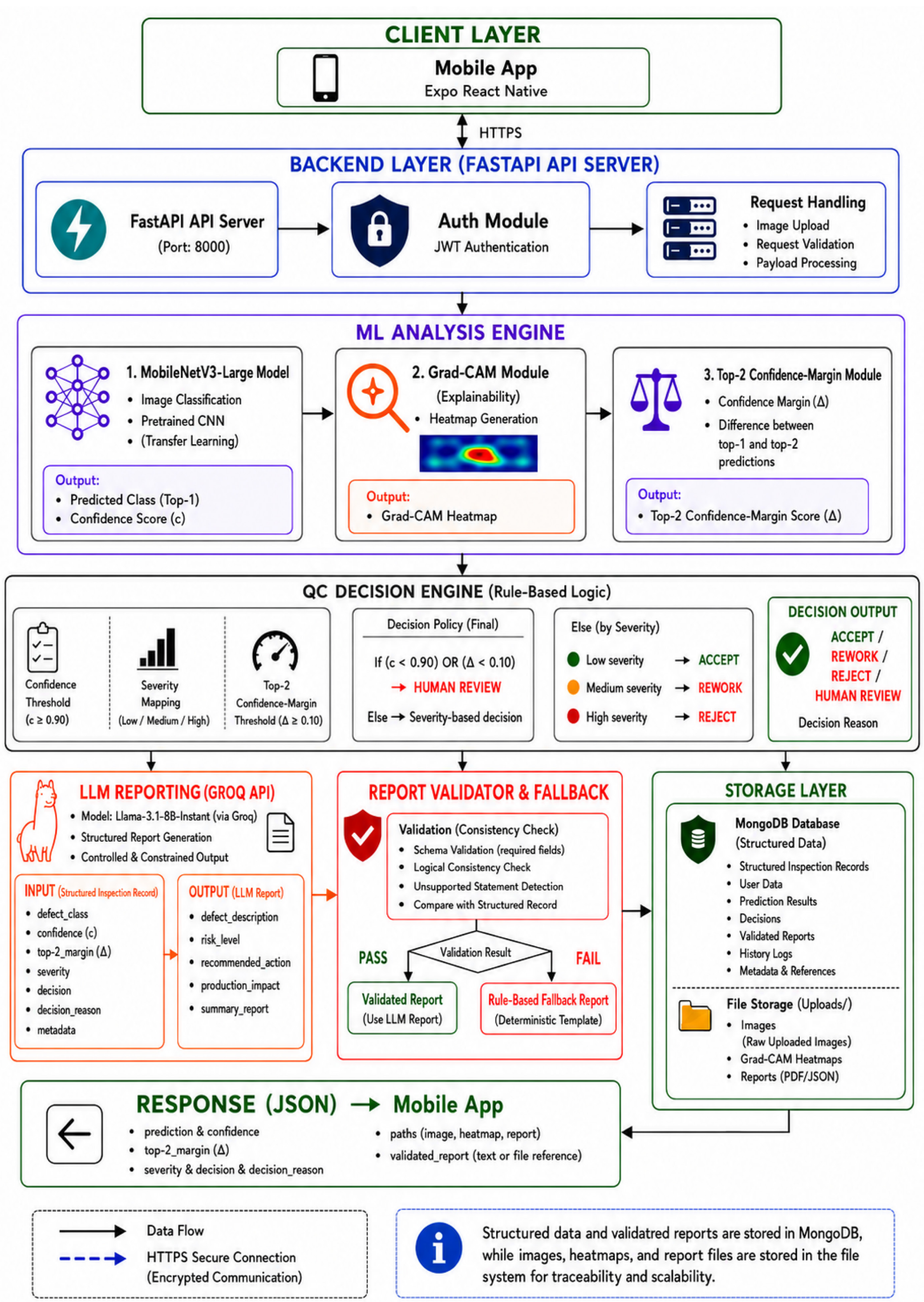

**Figure 10.** Overall architecture of the implemented RobustDefect-LLM mobile inspection and reporting system.

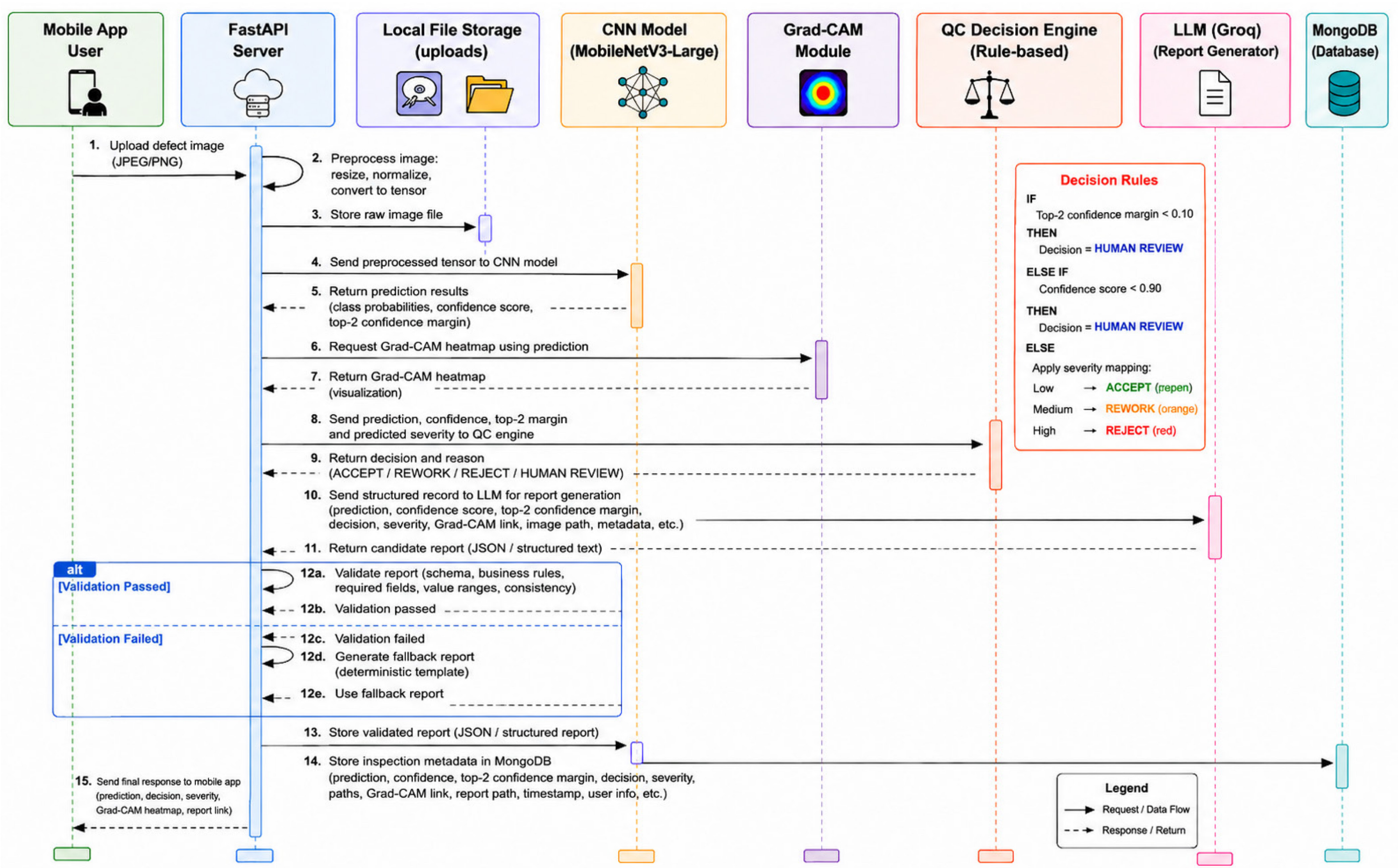

**Figure 11.** UML sequence diagram showing interactions among the components of RobustDefect-LLM.

## 10. Ethical, Practical, and Deployment Considerations

Industrial AI systems require more than high test accuracy. They must also be transparent, auditable, safe, and aligned with human responsibility. RobustDefect-LLM uses Grad-CAM to improve transparency, stores inspection history for traceability, and uses decision rules that can trigger human review when confidence is insufficient. However, the model should not be used as a fully autonomous quality-control authority without operator supervision.

Data privacy risks are limited because the dataset contains industrial surface images rather than personal data. Nevertheless, if the system is deployed in a factory, image metadata, operator accounts, logs, and production identifiers must be protected according to relevant privacy and security requirements, including GDPR and KVKK where applicable [30], [31]. Access control, secure API communication, and audit logging should be added before production deployment.

The most important technical limitations are dataset size, lack of a no-defect class, sensitivity to noise, and limited real-world validation. The current prototype runs locally and is not optimized for concurrent users or hardware-level integration with production lines. Future deployment should include camera calibration, input quality validation, edge or cloud inference optimization, and systematic testing under real lighting, motion, and material variations. The principal ethical, practical, and deployment limitations of the proposed framework, together with their recommended mitigation strategies, are summarized in Table 15.

**Table 15.** Ethical and practical limitations with recommended mitigation strategies.

| Category | Issue | Impact | Mitigation |
|---|---|---|---|
| **Transparency** | CNN decisions remain partly black-box | operators may not fully trust outputs | use Grad-CAM and structured reports |
| **Technical robustness** | noise and poor image quality reduce accuracy | risk of false decisions | robust training and input quality checks |
| **Safety** | false positives and false negatives | financial loss or quality escape | human verification for uncertain cases |
| **Data limitations** | small dataset and no no-defect class | limited generalization | expand dataset and add normal class |
| **Deployment** | local prototype only | limited scalability | cloud/edge deployment and load testing |
| **Security** | stored images and metadata require protection | possible leakage of production data | authentication, encryption, and audit logs |

## 11. Discussion

MobileNetV3-Large and DenseNet121 both achieved 100% maximum validation accuracy. Because the two models exhibited identical validation performance, MobileNetV3-Large was selected as the primary deployment model before held-out in-domain test evaluation based on its lower computational complexity, smaller deployment footprint, and shorter training time. The held-out in-domain test partition was then used exclusively for the final performance evaluation. Under this evaluation, MobileNetV3-Large achieved the numerically highest test accuracy (99.26%), followed closely by DenseNet121 (98.89%), EfficientNet-B0 (95.19%), and ResNet50 (79.63%). Although MobileNetV3-Large achieved the highest observed accuracy, the exact two-sided McNemar test showed that its performance difference relative to DenseNet121 was not statistically significant. These findings indicate that MobileNetV3-Large and DenseNet121 achieved statistically comparable classification performance on the held-out in-domain test partition, while MobileNetV3-Large provided a more favorable balance between predictive performance and computational efficiency.

Beyond classification performance, the proposed RobustDefect-LLM framework integrates Grad-CAM explainability, top-2 confidence margin analysis, rule-based quality-control decisions, AI-assisted reporting, and mobile deployment into a single inspection workflow. Table 16 compares the proposed framework with representative studies and highlights its end-to-end system integration rather than focusing solely on classification performance.

**Table 16.** Functional comparison of RobustDefect-LLM with selected industrial defect inspection approaches.

| Function | Damacharla et al. [18] | Chen et al. [10] | He et al. [1] | Shukla et al. [12] | RobustDefect-LLM |
|---|---|---|---|---|---|
| **Dataset** | Severstal steel defect dataset | NEU-DET and PV datasets | Multiple industrial datasets | MVTec AD and industrial datasets | NEU-DET surface defect dataset |
| **Main architecture / scope** | TLU-Net: U-Net with transfer-learning encoders | Deep-learning-based defect detection | Survey of deep-learning inspection methods | Review of deep anomaly-detection methods | MobileNetV3-Large with transfer learning |
| **Task type** | Steel defect classification and segmentation | Industrial component defect detection | Surface defect inspection review | Industrial anomaly-detection review | Industrial surface defect classification |
| **Explainability support** | Not a primary component | Not a primary component | Partial coverage in surveyed literature | Partial coverage in reviewed literature | Grad-CAM visualization |
| **Uncertainty analysis** | Not integrated | Not integrated as an operational layer | Not an integrated system layer | Discussed within anomaly-detection context | top-2 confidence margin analysis and low-confidence review logic |
| **Automated QC decision** | Not integrated | Not integrated | Not applicable to survey | Not applicable to review | HUMAN REVIEW / ACCEPT / REWORK / REJECT rule layer |
| **AI-assisted reporting** | Not integrated | Not integrated | Not applicable to survey | Not applicable to review | Controlled LLM-assisted inspection reports |
| **Robustness evaluation** | Transfer-learning and data-regime analysis | Partial robustness-oriented evaluation | Robustness challenges covered across literature | Robustness and generalization central to review | Noise/stress testing and qualitative error analysis |
| **Mobile / Web integration** | Not reported as core contribution | Not reported as core contribution | Not applicable to survey | Not applicable to review | React Native/Expo client with FastAPI backend and persistent history |

The comparison of RobustDefect-LLM with selected industrial defect inspection approaches presented in Table 16 highlights an important distinction between model-level innovation and system-level integration. Damacharla et al. [18] focus on transfer learning for steel defect classification and segmentation, demonstrating how pretrained encoders can improve data efficiency. Chen et al. [10] represent the deep-learning-based defect detection line, where the central objective is accurate identification and localization of industrial defects. The survey by He et al. [1] and the anomaly-detection review by Shukla et al. [12] demonstrate the breadth of modern industrial inspection methods and the continuing importance of robustness and generalization. RobustDefect-LLM does not seek to replace these model families; rather, it combines a compact supervised classifier with explainability, confidence-aware decision support, controlled reporting, traceability, and mobile access within a single demonstrable inspection workflow.

The principal contribution of RobustDefect-LLM is therefore the integration of six elements that are often evaluated separately: (1) efficient transfer-learned defect classification, (2) Grad-CAM-based visual evidence, (3) confidence and top-2 confidence margin analysis, (4) rule-based conversion of model outputs into HUMAN REVIEW, ACCEPT, REWORK, or REJECT pathways, (5) controlled AI-assisted reporting constrained by structured prediction fields, and (6) a full-stack inspection prototype with persistent history and mobile visualization. The experimental contribution complements this architecture through a baseline comparison, robustness testing, confusion-matrix analysis, and qualitative failure analysis. Together, these elements make the work an applied AI and software-engineering contribution centered on trustworthy workflow integration rather than a claim of a fundamentally new classifier or language model.

The bootstrap confidence interval analysis further supports the quantitative evaluation by showing that MobileNetV3-Large and DenseNet121 maintain relatively narrow confidence intervals on the held-out in-domain test partition. These intervals characterize the precision of the observed test metrics with respect to resampling of the current test data. However, they do not capture variability arising from model initialization, data-order changes, or independent training runs.

The paired McNemar comparison contained five discordant test samples, with three favoring MobileNetV3-Large and two favoring DenseNet121, resulting in no statistically significant difference ($p = 1.000$). MobileNetV3-Large had already been selected during validation because it matched DenseNet121's maximum validation accuracy while requiring fewer FLOPs, a smaller deployment footprint, and a shorter training time. Its slightly higher held-out test accuracy was observed only after model selection and was not statistically significant.

The results also expose clear limitations. Performance decreases under noisy or low-quality inputs, several visually similar defect classes remain confusable, the dataset is relatively small, and no explicit normal or no-defect class is included. The current robustness evidence is therefore informative but not sufficient to establish factory-level reliability. Likewise, Grad-CAM improves transparency but does not provide causal explanations, and the LLM-generated report should remain subordinate to structured model outputs and human review. These findings suggest that practical deployment should include input-quality checks, production-calibrated confidence and margin thresholds [19], a reject-or-review pathway for ambiguous cases, broader multi-site data, and prospective validation under real production conditions.

From an operational perspective, the prototype demonstrates how model outputs can be translated into an inspection process that is easier to review and audit. The mobile interface presents the source image, predicted class, confidence, severity, Grad-CAM evidence, decision result, report, and historical records in a unified interaction flow. This is the key practical distinction of the system: its value lies not only in achieving high benchmark performance, but in connecting perception to explanation and action while preserving a human review point for uncertain or safety-critical decisions.

## 12. Conclusion

This paper presented RobustDefect-LLM, an integrated framework for explainable industrial surface defect classification with AI-assisted reporting. The proposed system combines MobileNetV3-Large, Grad-CAM visualization, confidence-aware quality-control decisions, structured LLM-based reporting, and a mobile inspection interface within a unified workflow. Experimental evaluation on the held-out in-domain test partition showed that MobileNetV3-Large achieved the numerically highest test accuracy (99.26%) among the evaluated models, followed closely by DenseNet121, EfficientNet-B0, and ResNet50.

The results demonstrate that combining defect classification with explainability, confidence analysis, and rule-based decision support provides a broader and more operationally interpretable inspection workflow than classification alone. Furthermore, under the controlled nominal evaluation conditions, all 100 generated reports passed the predefined deterministic validation checks; however, failure-injection and adverse-condition testing remain necessary before production deployment. The developed prototype further demonstrates that these capabilities can be integrated into a mobile-oriented inspection workflow with inspection history, visualization, and AI-assisted reporting.

Bootstrap-based confidence interval analysis characterized the test-sample uncertainty of the reported performance, while the exact McNemar test showed no statistically significant difference between MobileNetV3-Large and DenseNet121 ($p = 1.000$). MobileNetV3-Large was retained as the primary deployment model because it had been selected during the validation stage based on validation performance and computational efficiency. On the held-out in-domain test partition, it subsequently achieved the numerically highest observed performance among the evaluated models.

Future work will focus on expanding the dataset, introducing a defect-free class, evaluating the framework under real industrial production conditions, improving robustness against challenging environments, and optimizing deployment for cloud and edge devices. Future studies will also investigate multimodal inspection, anomaly detection, segmentation, uncertainty-aware deep learning models, and more comprehensive evaluation of controlled AI-assisted reporting under real industrial operating conditions to further improve the practicality and reliability of industrial inspection systems.

## Code and Data Availability

The source code developed for the RobustDefect-LLM framework, including model training, evaluation, robustness testing, confidence analysis, Grad-CAM generation, CPU inference benchmarking, decision-support logic, and controlled LLM-report validation scripts, is publicly available through the project repositories [32], [33]. The backend and deep-learning implementation are available at [32], while the React Native mobile application is available at [33]. The repositories include the documentation, citation metadata, software license, and evaluation scripts required to support reproducibility. The NEU-DET dataset is publicly available from Northeastern University and is therefore not redistributed in the repositories.

## References

[1] Y. He, S. Li, X. Wen, and J. Xu, "A Survey on Surface Defect Inspection Based on Generative Models in Manufacturing," *Applied Sciences*, vol. 14, no. 15, Art. no. 6774, 2024, doi: 10.3390/app14156774.

[2] R. Ameri, C.-C. Hsu, and S. S. Band, "A systematic review of deep learning approaches for surface defect detection in industrial applications," *Engineering Applications of Artificial Intelligence*, vol. 130, Art. no. 107717, 2024, doi: 10.1016/j.engappai.2023.107717.

[3] Y. Liu, C. Zhang, and X. Dong, "A survey of real-time surface defect inspection methods based on deep learning," *Artificial Intelligence Review*, vol. 56, no. 10, pp. 12131–12170, 2023, doi: 10.1007/s10462-023-10475-7.

[4] M. Prunella, R. M. Scardigno, D. Buongiorno, A. Brunetti, N. Longo, R. Carli, M. Dotoli, and V. Bevilacqua, "Deep Learning for Automatic Vision-Based Recognition of Industrial Surface Defects: A Survey," *IEEE Access*, vol. 11, pp. 43370–43423, 2023, doi: 10.1109/ACCESS.2023.3271748.

[5] A. Krizhevsky, I. Sutskever, and G. E. Hinton, "ImageNet classification with deep convolutional neural networks," *Communications of the ACM*, vol. 60, no. 6, pp. 84–90, 2017, doi: 10.1145/3065386.

[6] Y. LeCun, Y. Bengio, and G. Hinton, "Deep learning," *Nature*, vol. 521, pp. 436–444, 2015, doi: 10.1038/nature14539.

[7] J. Deng, W. Dong, R. Socher, L.-J. Li, K. Li, and L. Fei-Fei, "ImageNet: A large-scale hierarchical image database," in *Proc. IEEE Conf. Computer Vision and Pattern Recognition (CVPR)*, 2009, pp. 248–255, doi: 10.1109/CVPR.2009.5206848.

[8] S. Ioffe and C. Szegedy, "Batch normalization: Accelerating deep network training by reducing internal covariate shift," in *Proc. 32nd Int. Conf. Machine Learning (ICML)*, vol. 37, 2015, pp. 448–456.

[9] F. Chollet, *Deep Learning with Python*, 2nd ed. Shelter Island, NY, USA: Manning Publications, 2021.

[10] Z. Chen, X. Feng, L. Liu, and Z. Jia, "Surface defect detection of industrial components based on vision," *Scientific Reports*, vol. 13, Art. no. 22136, 2023, doi: 10.1038/s41598-023-49359-9.

[11] K. Song and Y. Yan, "A noise robust method based on completed local binary patterns for hot-rolled steel strip surface defects," *Applied Surface Science*, vol. 285, pp. 858–864, 2013, doi: 10.1016/j.apsusc.2013.09.002.

[12] V. Shukla, A. Shukla, S. P. S. K., and S. Shukla, "A systematic survey: role of deep learning-based image anomaly detection in industrial inspection contexts," *Frontiers in Robotics and AI*, vol. 12, Art. no. 1554196, 2025, doi: 10.3389/frobt.2025.1554196.

[13] R. R. Selvaraju, M. Cogswell, A. Das, R. Vedantam, D. Parikh, and D. Batra, "Grad-CAM: Visual explanations from deep networks via gradient-based localization," in *Proc. IEEE Int. Conf. Computer Vision (ICCV)*, 2017, pp. 618–626, doi: 10.1109/ICCV.2017.74.

[14] A. Grattafiori et al., "The Llama 3 Herd of Models," *arXiv preprint arXiv:2407.21783*, 2024, doi: 10.48550/arXiv.2407.21783.

[15] Groq, "Groq API Documentation." [Online]. Available: https://console.groq.com/docs. Accessed: Jul. 2026.

[16] K. He, X. Zhang, S. Ren, and J. Sun, "Deep residual learning for image recognition," in *Proc. IEEE Conf. Computer Vision and Pattern Recognition (CVPR)*, 2016, pp. 770–778, doi: 10.1109/CVPR.2016.90.

[17] M. Tan and Q. V. Le, "EfficientNet: Rethinking model scaling for convolutional neural networks," in *Proc. 36th Int. Conf. Machine Learning (ICML)*, vol. 97, 2019, pp. 6105–6114.

[18] P. Damacharla, A. Rao M. V., J. Ringenberg, and A. Y. Javaid, "TLU-Net: A Deep Learning Approach for Automatic Steel Surface Defect Detection," in *Proc. 2021 Int. Conf. Applied Artificial Intelligence (ICAPAI)*, 2021, pp. 1–6, doi: 10.1109/ICAPAI49758.2021.9462060.

[19] C. Guo, G. Pleiss, Y. Sun, and K. Q. Weinberger, "On calibration of modern neural networks," in *Proc. 34th Int. Conf. Machine Learning (ICML)*, vol. 70, 2017, pp. 1321–1330.

[20] Y. Geifman and R. El-Yaniv, "Selective classification for deep neural networks," in *Advances in Neural Information Processing Systems 30 (NeurIPS)*, 2017, pp. 4878–4887.

[21] G. Huang, Z. Liu, L. van der Maaten, and K. Q. Weinberger, "Densely Connected Convolutional Networks," in *Proc. IEEE Conf. Computer Vision and Pattern Recognition (CVPR)*, 2017, pp. 4700–4708, doi: 10.1109/CVPR.2017.243.

[22] A. Howard, M. Sandler, G. Chu, L.-C. Chen, B. Chen, M. Tan, W. Wang, Y. Zhu, R. Pang, V. Vasudevan, Q. V. Le, and H. Adam, "Searching for MobileNetV3," in *Proc. IEEE/CVF Int. Conf. Computer Vision (ICCV)*, 2019, pp. 1314–1324, doi: 10.1109/ICCV.2019.00140.

[23] D. P. Kingma and J. Ba, "Adam: A method for stochastic optimization," in *Proc. Int. Conf. Learning Representations (ICLR)*, 2015.

[24] I. Loshchilov and F. Hutter, "Decoupled weight decay regularization," in *Proc. Int. Conf. Learning Representations (ICLR)*, 2019.

[25] L. Yu, "THOP: PyTorch-OpCounter," GitHub repository. [Online]. Available: https://github.com/Lyken17/pytorch-OpCounter. Accessed: Jul. 2026.

[26] A. Paszke et al., "PyTorch: An imperative style, high-performance deep learning library," in *Advances in Neural Information Processing Systems 32 (NeurIPS)*, 2019, pp. 8024–8035.

[27] FastAPI, "FastAPI Documentation." [Online]. Available: https://fastapi.tiangolo.com/. Accessed: Jul. 2026.

[28] MongoDB, "MongoDB Documentation." [Online]. Available: https://www.mongodb.com/docs/. Accessed: Jul. 2026.

[29] Expo, "Expo Documentation." [Online]. Available: https://docs.expo.dev/. Accessed: Jul. 2026.

[30] European Parliament and Council of the European Union, "Regulation (EU) 2016/679—General Data Protection Regulation (GDPR)," *Official Journal of the European Union*, vol. L 119, pp. 1–88, May 4, 2016.

[31] Republic of Türkiye, "Law No. 6698 on the Protection of Personal Data (Kişisel Verilerin Korunması Kanunu—KVKK)," *Official Gazette*, no. 29677, Apr. 7, 2016.

[32] N. Düşünmez, "RobustDefect-LLM Backend Repository," GitHub. [Online]. Available: https://github.com/nazlican530/industrial-defect-qc. Accessed: August 2026.

[33] N. Düşünmez, "RobustDefect-LLM Mobile Application," GitHub. [Online]. Available: https://github.com/nazlican530/mobile-app. Accessed: August 2026.